\documentclass[10pt,twocolumn,letterpaper]{article}

\usepackage[pagenumbers]{cvpr} % To force page numbers, e.g. for an arXiv version

\usepackage[table]{xcolor}
\usepackage{cuted}
\usepackage{capt-of}
\usepackage[accsupp]{axessibility}  % Improves PDF readability for those with disabilities.
\definecolor{p_po}{RGB}{255,192,0}
\definecolor{p_ib}{RGB}{146,208,80}
\definecolor{p_eb}{RGB}{0,176,240}
\definecolor{epsilon_cig}{RGB}{208,206,206}
\definecolor{phi_cn}{RGB}{255,192,0}
\definecolor{u_iv}{RGB}{112,48,160}
\definecolor{phi_rf}{RGB}{233,113,50}

\definecolor{cvprblue}{rgb}{0.21,0.49,0.74}
\usepackage[pagebackref,breaklinks,colorlinks,allcolors=cvprblue]{hyperref}

\def\paperID{} % *** Enter the Paper ID here
\def\confName{CVPR}
\def\confYear{2026}

\title{PHAC: Promptable Human Amodal Completion}

\author{Seung Young Noh \qquad Ju Yong Chang\\
Kwangwoon University\\
{\tt\small \{kelvinnoh,jychang\}@kw.ac.kr}
}

\begin{document}
% \maketitle
\twocolumn[\maketitle\vspace{-2mm}\begin{center}
    \vspace{-2.5em}
    \includegraphics[width=0.9\linewidth]{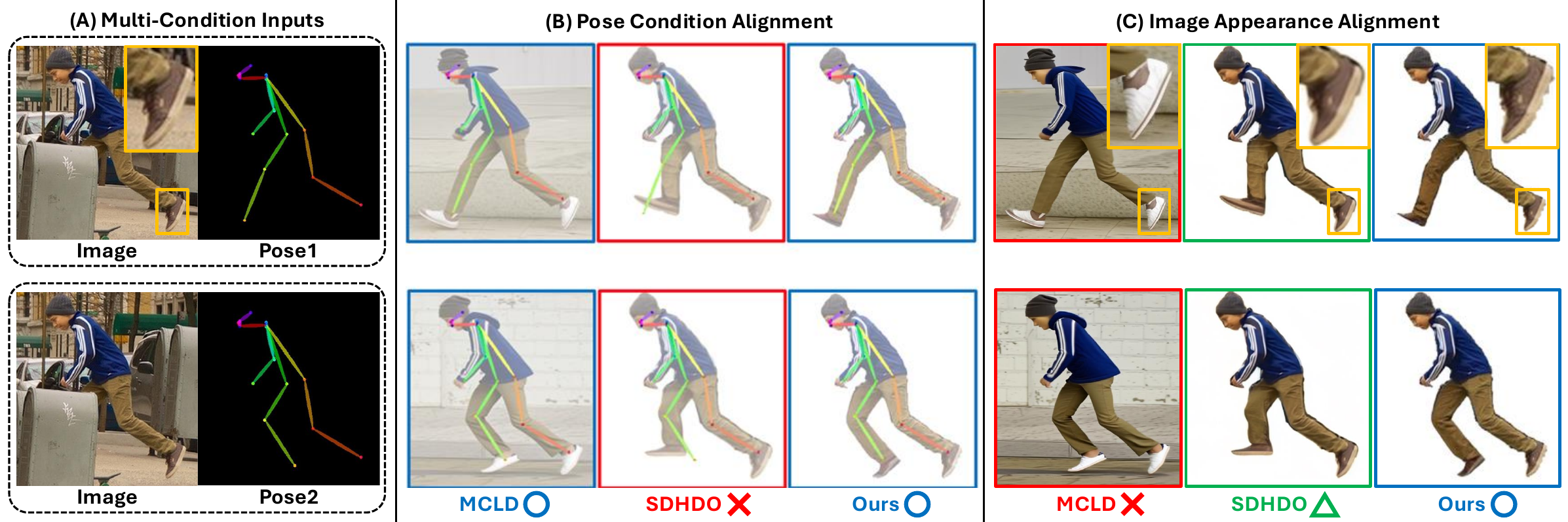}
    \vspace{-2mm}
    \captionof{figure}{
    Assume that, as in (A), a single input image is paired with multiple target poses. The desired behavior is that each generated image aligns with its target pose while preserving the visible appearance. However, prior human amodal completion (HAC) methods such as SDHDO~\cite{noh2025sdhdo} often fail to align with the specified pose, as shown in (B), despite leveraging pose information. Pose-guided person image synthesis (PGPIS) methods such as MCLD~\cite{liu2025mcld} align well with the pose condition but degrade the visible appearance of the input, especially around the shoes, as shown in (C). While SDHDO preserves the visible appearance better than MCLD, its results remain blurry and show noticeable degradation. In contrast, given the multi-condition inputs in (A), our method simultaneously aligns with the target pose and preserves the visible appearance, producing the user-intended human image.
    }
    \label{fig1:teaser}
\end{center}
\vspace{-6mm}
\vspace{2mm}\bigbreak]
\begin{abstract}
Conditional image generation methods are increasingly used in human-centric applications, yet existing human amodal completion (HAC) models offer users limited control over the completed content. Given an occluded person image, they hallucinate invisible regions while preserving visible ones, but cannot reliably incorporate user-specified constraints such as a desired pose or spatial extent. As a result, users often resort to repeatedly sampling the model until they obtain a satisfactory output. Pose-guided person image synthesis (PGPIS) methods allow explicit pose conditioning, but frequently fail to preserve the instance-specific visible appearance and tend to be biased toward the training distribution, even when built on strong diffusion model priors. To address these limitations, we introduce promptable human amodal completion (PHAC), a new task that completes occluded human images while satisfying both visible appearance constraints and multiple user prompts. Users provide simple point-based prompts, such as additional joints for the target pose or bounding boxes for desired regions; these prompts are encoded using ControlNet modules specialized for each prompt type. These modules inject the prompt signals into a pre-trained diffusion model, and we fine-tune only the cross-attention blocks to obtain strong prompt alignment without degrading the underlying generative prior. To further preserve visible content, we propose an inpainting-based refinement module that starts from a slightly noised coarse completion, faithfully preserves the visible regions, and ensures seamless blending at occlusion boundaries. Extensive experiments on the HAC and PGPIS benchmarks show that our approach yields more physically plausible and higher-quality completions, while significantly improving prompt alignment compared with existing amodal completion and pose-guided synthesis methods.
\end{abstract}
    
\section{Introduction}
\label{sec1:intro}

%% Condition image generation / importance of promptable and alignment
Image generation has been widely adopted in recent years and continues to attract strong interest. In particular, conditional image generation, which synthesizes images that satisfy user-specified conditions, has attracted increasing attention relative to unconditional generation, which produces diverse images without explicit guidance. For conditional image generation, beyond output quality and realism, it is crucial that models be easily controlled with diverse user prompts and that the generated results faithfully align with these prompts. In practice, users frequently refine and share text prompts to better express their intended conditions, and frameworks such as ControlNet~\cite{zhang2023controlnet} are widely used to handle diverse conditioning inputs.

%% HAC / limitation of HAC
Human amodal completion (HAC) takes an image of an occluded human as input and synthesizes a plausible human shape and appearance in the occluded regions while preserving the visible content. Therefore, HAC can be viewed as a form of conditional image generation. Producing a complete human image from an incomplete input enables downstream applications such as novel-view synthesis, 3D avatar generation, and user-guided photo editing. For these reasons, amodal completion, including HAC, remains an active research topic. However, existing approaches~\cite{zhan2020sssd,ehsani2018segan,meng2022hand_do,zhou2021human_do,yan2019vehicle_do,papadopoulos2019pizza_do,liu2024ols_do,ozguroglu2024pix2gestalt,xu2024pdmc,noh2025sdhdo} generally cannot incorporate user-specified prompts as conditioning signals. For example, when a user provides a desired pose for an occluded person, there is no straightforward way to enforce that pose during generation. As a result, users may need to run the model multiple times to obtain a satisfactory result, and models often produce implausible human completions in the process. Recent works~\cite{ozguroglu2024pix2gestalt,noh2025sdhdo} can inject pose conditions using pre-trained ControlNet modules~\cite{zhang2023controlnet}. However the generated images often remain poorly aligned with the given pose constraint, as shown in Fig.~\ref{fig1:teaser}(B).

%% PGPIS / limitation of PGPIS
A promising direction for addressing these limitations is pose-guided person image synthesis (PGPIS). PGPIS takes a source image, a source pose, and a target pose as input, and synthesizes an image that aligns with the target pose while remaining consistent with the source appearance. In this regard, PGPIS can be viewed as conditional image generation, where the source image provides appearance information and the target pose serves as the pose condition. In the HAC setting, the target pose contains all visible joints from the source pose while adding only the invisible ones. The target image should preserve the visible regions while satisfying this extended pose condition. Despite leveraging the strong priors of pre-trained diffusion models (DMs)~\cite{ho2020ddpm,rombach2022sd}, existing PGPIS methods~\cite{bhunia2023pidm,han2023pocold,lu2024cfld,liu2025mcld} often struggle to maintain instance-specific visible appearance and instead tend to produce content reflecting the training distribution, as illustrated in Fig.~\ref{fig1:teaser}(C).

%% common limitation of amodal completion and PGPIS
Both amodal completion and PGPIS require synthesizing plausible human appearance in invisible regions based on the visible context. Recent approaches improve performance by leveraging DM priors or fine-tuning the models. However, latent space denoising tends to discard fine-grained visible details. To mitigate this issue, previous work has fine-tuned decoders to reduce information loss in visible regions~\cite{noh2025sdhdo,avrahami2023bld} or has used UV coordinate-based texture maps~\cite{guler2018densepose} to preserve appearance~\cite{liu2025mcld,han2023pocold}. Nevertheless, decoder fine-tuning may introduce blurry outputs and blending artifacts at mask boundaries, while texture map approaches often lose details when UV coordinates are noisy.

%% proposed method
To overcome these limitations, we introduce a new task called \emph{promptable human amodal completion (PHAC)}. The goal is to generate human images that satisfy both the visible appearance constraints and multiple user-specified prompts, in which users can specify prompts using only a few points. For pose prompts, the user specifies additional joints for the desired pose given the visible ones. For bounding-box (bbox) prompts, the user simply selects two points. We treat these simple prompts, together with the occluded input image, as conditioning signals for HAC. Consequently, unlike prior HAC~\cite{zhou2021human_do,noh2025sdhdo} and general amodal completion~\cite{ozguroglu2024pix2gestalt,xu2024pdmc,zhan2020sssd,liu2024ols_do} methods, our generated images can reflect user-specified constraints. Beyond steering generation toward user intention, prompts also provide auxiliary guidance for the invisible regions, which enables the diffusion model to produce more physically plausible and higher-quality results. The prompts are injected into the denoising U-Net via ControlNet modules specialized for each prompt type. To preserve the rich priors of pre-trained DMs while achieving strong prompt alignment, we fine-tune only the cross-attention blocks~\cite{ho2024sith}.

%% refinement module
While the generated human image aligns with the user prompts, the visible appearance can still degrade during latent space denoising and variational autoencoder (VAE)~\cite{kingma2013vae} reconstruction. To address this issue, we refine the generated image using an inpainting model~\cite{podell2023sdxl}. Standard inpainting models synthesize pixels within a masked region and blend them into the surrounding context. By instead using a model that preserves unmasked regions and seamlessly fuses the synthesized content, our refinement step maintains the visible appearance and yields a coherent final image. The refinement module takes the coarse completion, the incomplete image, and the invisible-region mask as inputs, preserves the visible regions exactly, and applies only a small amount of denoising to ensure smooth blending. Denoising from a low noise level, rather than from pure random noise, further helps retain both visible and generated details and preserves boundary consistency.

%% contributions
The key contributions of this work are as follows:
\begin{itemize}
\item We introduce the new task of promptable human amodal completion, which completes occluded human images while satisfying both visible appearance constraints and user-specified prompts. We also present the first framework to address this task.
\item We propose an inpainting-based refinement module that preserves the visible appearance and synthesized content while maintaining boundary continuity. This module can also serve as a plug-and-play component for improving the outputs of other diffusion models.
\item We demonstrate, through extensive quantitative and qualitative evaluations, that our approach achieves superior prompt alignment and produces more physically plausible and higher-quality images than existing amodal completion and PGPIS methods.
\end{itemize}

\section{Related Work}
\label{sec2:related}

%% Sec 2.1
\subsection{Conditional Image Generation}

Conditional image generation aims to synthesize images that satisfy specified conditions. Early work explored conditioning mechanisms in VAEs~\cite{sohn2015cvae,bao2017cvae_gan} and generative adversarial networks (GANs)~\cite{mirza2014cgan,goodfellow2014gan}. Subsequent GAN-based methods extended conditioning to image-to-image translation~\cite{isola2017pix2pix,zhu2017cyclegan} and text-guided image generation~\cite{zhang2017stackgan,xu2018attngan}. DMs~\cite{ho2020ddpm,rombach2022sd} further advanced conditional synthesis, particularly in large-scale text-to-image generation~\cite{ramesh2021dalle,nichol2021glide,saharia2022imagen}. However, text prompts alone often fail to express fine spatial constraints such as geometry or human pose. ControlNet~\cite{zhang2023controlnet} addresses this limitation by injecting image-based structural cues (e.g., 2D poses and normal maps) into a pre-trained DM. While ControlNet improves spatial controllability, faithfully preserving the detailed appearance of a specific person remains challenging.

%% Sec 2.2
\subsection{Amodal Completion}

Amodal completion aims to recover both the amodal region and RGB content of occluded areas. Early work predicted only the amodal region, using bounding boxes~\cite{kar2015amodal_bbox1,hsieh2023amodal_bbox2} or amodal masks~\cite{li2016amodal_seg1,zheng2021vinv_seg2,follmann2019amodal_seg4,zhu2017amodal_seg5,qi2019amodal_seg6,dhamo2019amodal_seg7,li2023muva_seg8,hu2019sail_seg9,reddy2022walt_seg10,vuong2024walt3d_seg11,ke2021amodal_seg12}, and subsequent studies inferred amodal masks directly from visible masks~\cite{zhan2024amodal_com1,nguyen2021amodal_com2}. Later two-stage approaches~\cite{zhan2020sssd,ehsani2018segan,yan2019vehicle_do,papadopoulos2019pizza_do} first reconstructed the occluded region and then synthesized its RGB appearance. Although effective within training distributions, these methods often struggle to generalize to complex or unseen occlusion patterns. To improve robustness, prior work restricted the problem to specific categories such as hands~\cite{meng2022hand_do} or full-body humans~\cite{zhou2021human_do,noh2025sdhdo}, and diffusion-based methods~\cite{ozguroglu2024pix2gestalt,xu2024pdmc} leveraged pre-trained generative priors to enhance zero-shot performance. However, existing methods do not support user-specified conditioning and may still lose fine visible details due to denoising.

%% Sec 2.3
\subsection{Pose-Guided Person Image Synthesis}

PGPIS methods synthesize images that align with a target pose
while preserving the source appearance. Early PGPIS approaches relied on GANs~\cite{ma2017pg2,siarohin2018defor_gan,dong2018sggan,zhang2021pise,lv2021spgnet}, while later methods improved pose-texture interaction using attention mechanisms~\cite{zhu2019patn} and transformer-based architectures~\cite{zhang2022dptn}. More recent diffusion-based approaches~\cite{bhunia2023pidm,han2023pocold,lu2024cfld,liu2025mcld} inject the source appearance and target pose into the latent space of pre-trained DMs. Nonetheless, these models are often biased toward training datasets such as DeepFashion~\cite{liu2016deepfashion}, and may synthesize appearance that reflects dataset-specific biases rather than preserving the instance-specific visual characteristics of zero-shot inputs.

%% Sec 2.4
\subsection{Prompts and Human Interaction}

User prompting provides an effective mechanism for steering model behavior when automatic inference does not align with user intention. In computer vision, SAM~\cite{kirillov2023sam} supports point- or box-based prompts to generate segmentation masks, while text-to-image DMs such as Stable Diffusion~\cite{rombach2022sd} accept text prompts and structural conditions via ControlNet~\cite{zhang2023controlnet}. Other recent works utilize spatial or semantic prompts for controllable generation, including semantic palettes~\cite{lee2025semanticdraw}, spatial cues for mesh or image generation~\cite{huang2025spar3d,jain2024peekaboo}, and auxiliary prompts that improve tasks such as human mesh recovery~\cite{wang2025prompthmr}, pose estimation~\cite{yang2023clickpose}, and 3D object detection~\cite{zhang2024inter_3dobj}. These studies demonstrate the effectiveness of lightweight prompting for guiding complex models. However, to the best of our knowledge, existing approaches neither address amodal completion with multiple prompt types nor integrate user-specified prompts with visible appearance constraints, which is the focus of our work.

\section{Proposed Method}
\label{sec3:method}

%===========================================================
%% Sec 3.1
\subsection{Preliminary: Conditional Latent Diffusion}
\label{sec3.1:preliminary}

DMs synthesize images by progressively denoising a sample that is initially drawn from random noise. Since performing this process directly in pixel space is computationally expensive and scales with image resolution, recent approaches adopt latent diffusion models, which operate in the latent space of a VAE. In this setting, the DM incorporates text embeddings and ControlNet-based conditioning signals into the denoising process.

For an input image $I$, the VAE encoder produces the clean latent $z_0 = \mathcal{E}(I)$. The noisy latent $z_t$ at timestep $t$ is obtained via the forward diffusion process:
\begin{equation}
    \label{eq1:z_t}
    z_t = \sqrt{\bar{\alpha}_t} z_0 + \sqrt{1 - \bar{\alpha}_t} \epsilon,
\end{equation}
where $\bar{\alpha}_t = \prod_{s=1}^{t} \alpha_s$ is defined by the diffusion schedule, and $\epsilon \sim \mathcal{N}(0, I)$.

To synthesize images aligned with the given conditions from $z_t$, the denoising U-Net $\epsilon_\theta$ and the ControlNet are trained with the following objective:
\begin{equation}
    \label{eq2:dm_loss}
    \mathcal{L}
    = \mathbb{E}_{z_0,\epsilon,t,c_\text{te},c_\text{pr}}
      \left[
        \big\| \epsilon - \epsilon_\theta(z_t, t, c_\text{te}, c_\text{pr}) \big\|_2^2
      \right],
\end{equation}
where $c_\text{te}$ denotes the text embedding extracted from a CLIP encoder~\cite{radford2021clip}, and $c_\text{pr}$ denotes the ControlNet-derived conditioning signal produced from the user-specified prompt.

%% Fig 2: user prompt
\begin{figure}
    \centering
    \includegraphics[width=0.8\linewidth]{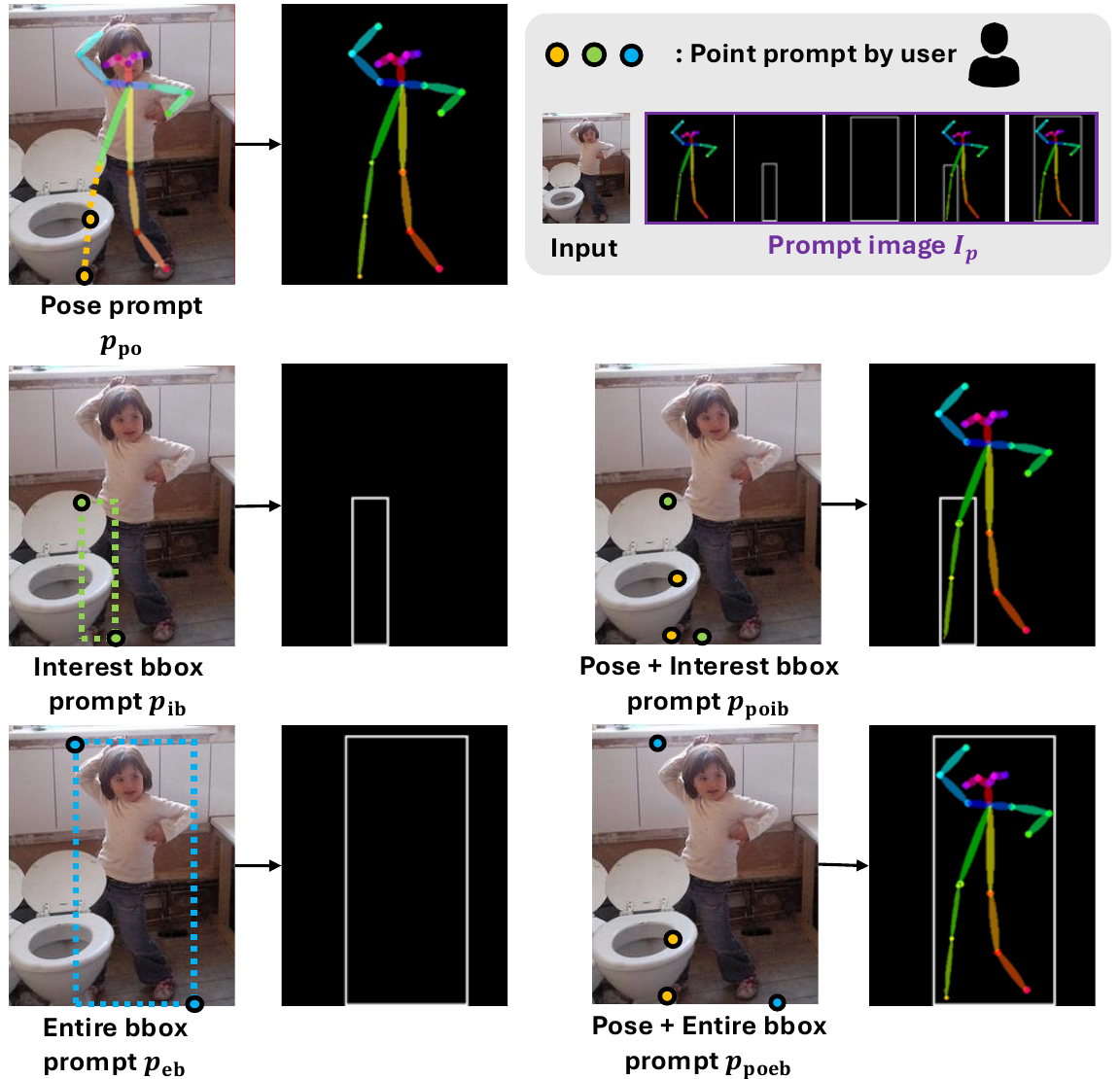}
    \vspace{-2mm}
    \caption{\textbf{User prompts}. Users specify the intended pose with point prompts, which we use to condition the model. For the pose prompt $p_\text{po}$, we use OpenPose~\cite{cao2019openpose} to detect the visible joints, show them to the user, who then adds the missing joints for the desired pose. Alternatively, the user selects two points to specify a bbox prompt, choosing either an interest-region bbox $p_\text{ib}$ or an entire-region bbox $p_\text{eb}$. To provide fine-grained control, the pose and bbox prompts can be combined, yielding $p_\text{poib}$ or $p_\text{poeb}$. To make effective use of the spatial information, we convert the point coordinates into a prompt image $I_\text{p}$ and use it as a conditioning input.}
    \label{fig2:prompt}
    \vspace{-4mm}
\end{figure}

%% Fig 3: method overview
\begin{figure*}
    \centering
    \includegraphics[width=0.9\linewidth]{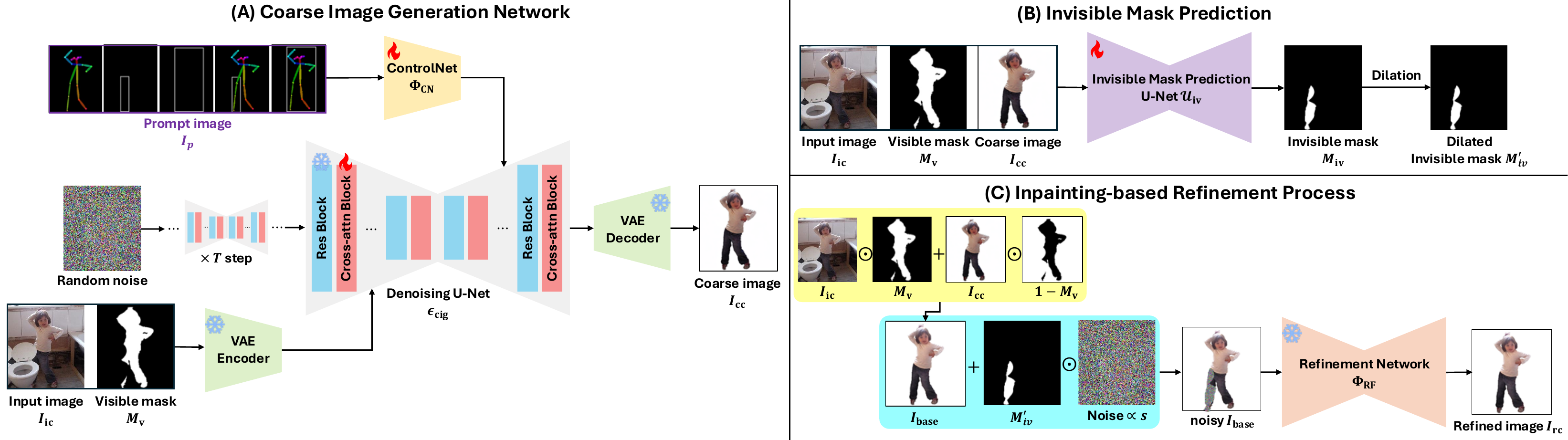}
    \vspace{-2mm}
    \caption{\textbf{Method overview}. Given an incomplete image $I_\text{ic}$ and a user prompt $P$, our PHAC framework processes them through (A) coarse image generation and (B, C) a refinement stage. In (A), the \textcolor{epsilon_cig}{denoising U-Net} $\epsilon_\text{cig}$ starts from random noise and denoises it for $T$ steps to generate a coarse complete image $I_\text{cc}$, conditioned on a prompt image $I_\text{p}$ (see Fig.~\ref{fig2:prompt}); $I_\text{p}$ is fed to a prompt-specific \textcolor{phi_cn}{ControlNet} $\Phi_\text{CN}$ to provide conditioning, and only the cross-attention blocks of $\epsilon_\text{cig}$ are fine-tuned to preserve the pre-trained prior. In (B), \textcolor{u_iv}{invisible mask prediction U-Net} $\mathcal{U}_\text{iv}$ predicts an invisible mask $M_\text{iv}$, which is then dilated to $M_\text{iv}^\prime$. In (C), we construct the base composite $I_\text{base}$ and add low-magnitude noise to the invisible region only. The \textcolor{phi_rf}{refinement network} $\Phi_\text{RF}$ then takes the noisy $I_\text{base}$ as input and outputs the refined completion $I_\text{rc}$, preserving the visible region while refining the coarse completion and mitigating boundary artifacts.}
    \label{fig3:method}
    \vspace{-4mm}
\end{figure*}

%===========================================================
%% Sec 3.2
\subsection{Promptable Human Amodal Completion}
\label{sec3.2:PHAC}

PHAC aims to synthesize a complete person image from an occluded input while preserving the input’s visible appearance and aligning with user-specified prompts. Prior amodal completion methods make different input assumptions: they either require an accurate visible-region mask as input~\cite{zhan2020sssd,zhou2021human_do,liu2024ols_do} or do not accept user-specified prompts~\cite{ozguroglu2024pix2gestalt,xu2024pdmc}. Although PGPIS takes as input a source image and a target pose, which are similar to our inputs, it is restricted to a single prompt type (e.g., pose map~\cite{bhunia2023pidm,lu2024cfld} or UV map~\cite{liu2025mcld,han2023pocold}). Our approach takes a user prompt as a conditioning input, including a pose prompt and a bbox prompt that specifies the region to synthesize. For a pose prompt, given the visible joints, the user only needs to add the additional joints for the desired pose; for a bbox prompt, the box can either specify only the interest region for synthesis or span the entire region, including visible regions. The pose and bbox prompts can also be combined. As illustrated in Fig.~\ref{fig2:prompt}, the user selects any of five prompt types and uses it as the user prompt $P$, defined as:
\begin{equation}
    \label{eq3:prompt_set}
    P \in \{ p_{\text{po}}, p_{\text{ib}}, p_{\text{eb}}, p_{\text{poib}}, p_{\text{poeb}} \},
\end{equation}
where $p_\text{po}$ is the pose prompt; $p_{\text{ib}}$ and $p_{\text{eb}}$ are the bbox prompts for the interest region and the entire region, respectively. $p_{\text{poib}}$ and $p_{\text{poeb}}$ denote composite prompts that combine a pose prompt with an interest-region bbox or an entire-region bbox, respectively.

The PHAC network $\Phi_\text{PHAC}$ takes an incomplete human image $I_\text{ic}$ and the user prompt $P$ as inputs and outputs a complete human image $I_\text{c}$ as follows:
\begin{equation}
    \label{eq4:phac}
    I_\text{c} = \Phi_\text{PHAC}(I_\text{ic}, P).
\end{equation}

The proposed method injects the prompt-based conditioning signal $c_\text{pr}$, which is computed by a ControlNet $\Phi_\text{CN}$ specialized for each prompt type, into the coarse image generation network $\Phi_\text{CIG}$. Conditioned on $c_\text{pr}$, $\Phi_\text{CIG}$ takes the incomplete image $I_\text{ic}$ and the visible mask $M_\text{v}$ as inputs and produces a coarse complete image $I_\text{cc}$, formulated as:
\begin{equation}
    \label{eq5:i_cc}
    c_\text{pr} = \Phi_\text{CN}(P),\;
    I_\text{cc} = \Phi_\text{CIG}(I_\text{ic}, M_\text{v}, c_\text{pr}).
\end{equation}

To mitigate degradation of the visible appearance arising from latent space denoising and VAE reconstruction, the refinement network $\Phi_\text{RF}$ takes as input the incomplete image $I_\text{ic}$, the coarse completion $I_\text{cc}$, the visible-region mask $M_\text{v}$, and the invisible-region mask $M_\text{iv}$, and outputs the refined complete image $I_\text{rc}$ as:
\begin{equation}
    \label{eq6:i_rc}
        I_\text{rc} = \Phi_\text{RF}(I_\text{ic}, I_\text{cc}, M_\text{v}, M_\text{iv}).
\end{equation}

The coarse image generation network and the refinement network are described in detail in Secs.~\ref{sec3.3:CIG} and~\ref{sec3.4:refine}, respectively.

%===========================================================
%% Sec 3.3
\subsection{Coarse Image Generation Network}
\label{sec3.3:CIG}

\noindent\textbf{User prompt.}
In PHAC, the user prompt encodes spatial information about the desired pose, and the goal is to synthesize an image conditioned on this information. Prompts that impose strong constraints on occluded regions, such as 3D cues (e.g., depth or normal maps) or an invisible mask, can serve as high-quality prompts. In practice, however, such prompts are difficult for users to create and provide to the network. To balance prompt effectiveness with usability, we instead define user-friendly point prompts that require only a small set of points on the image. To obtain the pose prompt $p_\text{po}$, we first detect the visible joints with an off-the-shelf 2D pose estimator~\cite{cao2019openpose}. Given the detected visible joints, the user adds the missing joints for the desired pose; these user-specified points form the pose prompt $p_\text{po}$ (\textcolor{p_po}{orange points} in Fig.~\ref{fig2:prompt}). For the bbox prompts $p_\text{ib}$ and $p_\text{eb}$, the user specifies two points to define either a bbox over the interest region (\textcolor{p_ib}{light-green points}) or a bbox covering the entire region, including visible areas (\textcolor{p_eb}{sky-blue points}). To provide a stronger constraint, the user can select a composite prompt that pairs a pose with a bbox, such as $p_\text{poib}$ or $p_\text{poeb}$.

\noindent\textbf{Coarse image generation.}
To effectively exploit the spatial information in the user prompt $P$, we convert it into a prompt image $I_\text{p}$ rather than using raw coordinate values.

The prompt image $I_\text{p}$ is fed to a prompt-specific ControlNet $\Phi_{\text{CN}}$ to produce the prompt-conditioning signal $c_\text{pr}$, computed as follows:
\begin{equation}
    \label{eq7:c_pr}
    c_\text{pr} = \Phi_\text{CN}(I_\text{p}).
\end{equation}

The text condition $c_\text{te}$ is obtained from the input image using the CLIP encoder~\cite{radford2021clip}. This step is omitted in Fig.~\ref{fig3:method} for simplicity. Unlike previous amodal completion methods~\cite{zhan2020sssd,zhou2021human_do,liu2024ols_do} that used ground-truth (GT) masks, we predict the visible mask $M_\text{v}$ from the input image using SAM~\cite{kirillov2023sam} as follows:
\begin{equation}
    \label{eq8:clip_sam}
    c_\text{te} = \text{CLIP}(I_\text{ic}),\;
    M_\text{v} = \text{SAM}(I_\text{ic}).
\end{equation}

Given the incomplete image $I_\text{ic}$, the visible mask $M_\text{v}$, the text condition $c_\text{te}$, and the prompt condition $c_\text{pr}$, the denoising U-Net $\epsilon_\text{cig}$ iteratively denoises a latent initialized from random noise to synthesize the coarse completion $I_\text{cc}$. The denoising U-Net $\epsilon_\text{cig}$ and the ControlNet $\Phi_\text{CN}$ are jointly trained to minimize the following objective:
\begin{equation}
    \label{eq9:cig_loss}
    \mathcal{L} = \mathbb{E}_{z_0, \epsilon, t, c_\text{te}, c_\text{pr}} \left[ \|\epsilon - \epsilon_\text{cig} (z_t, t, c_\text{te}, c_\text{pr}) \|_2^2 \right],
\end{equation}
where $t$ denotes the diffusion timestep with $0 \le t < 1000$. The clean latent $z_0$ from the input image $I_\text{ic}$ and the visible mask $M_\text{v}$ is obtained by the VAE encoder $\mathcal{E}$ as:
\begin{equation}
    \label{eq10:z_0}
    z_0 = \mathcal{E}(I_\text{ic})\oplus\mathcal{E}(M_\text{v}),
\end{equation}
where $\oplus$ denotes the concatenation operation.

To better align the generated image with the ControlNet-derived prompt condition $c_\text{pr}$, we fine-tune only the cross-attention blocks of the denoising U-Net $\epsilon_\text{cig}$, keeping the remaining weights frozen, as in~\cite{ho2024sith}. This enables $\epsilon_\text{cig}$ to generate images that align more effectively with the user prompt $P$. The remaining blocks are kept frozen to preserve the pre-trained DM prior, yielding plausible human appearance.

% Table 1: main table
\begin{table*}[t]
    \centering
    {\scriptsize
    \renewcommand{\arraystretch}{1.2}
    \setlength{\tabcolsep}{3pt}
    %\resizebox{\columnwidth}{!}{
    \begin{tabular}{l | c | cccccc | cccccc}
        \toprule
        \raisebox{-2ex}{Method} & \raisebox{-2ex}{User Prompt}
        & \multicolumn{6}{c|}{OccThuman2.0 test dataset (synthetic)}
        & \multicolumn{6}{c}{AHP test dataset (real)} \\
        \cline{3-8} \cline{9-14}
        &
        & LPIPS* $\downarrow$ & SSIM $\uparrow$ & KID* $\downarrow$ & MSE* $\downarrow$ & PSNR $\uparrow$ & Joint Err. $\downarrow$
        & LPIPS* $\downarrow$ & SSIM $\uparrow$ & KID* $\downarrow$ & MSE* $\downarrow$ & PSNR $\uparrow$ & Joint Err. $\downarrow$ \\
        \midrule
        PIDM~\cite{bhunia2023pidm} & 2D pose map & 126.33 & 0.797 & 56.91 & 27.34 & 16.80 & 113.72 & 143.15 & 0.804 & 23.71 & 25.33 & 16.61 & 40.47 \\
        MCLD~\cite{liu2025mcld} & UV map & 115.90 & 0.833 & 41.11 & 18.94 & 18.37 & 53.38 & 121.80 & 0.853 & 21.90 & 14.85 & 18.97 & 11.53 \\
        pix2gestalt~\cite{ozguroglu2024pix2gestalt} & - & 90.11 & 0.911 & 16.51 & 7.58 & 22.63 & 36.65 & 75.73 & 0.942 & \underline{5.98} & 5.22 & 24.06 & 10.96 \\
        pix2gestalt~\cite{ozguroglu2024pix2gestalt}\dag & 2D pose map & 88.58 & 0.914 & 16.75 & \underline{6.93} & 22.94 & \underline{31.37} & 75.25 & 0.943 & 6.35 & \underline{4.87} & 24.25 & 10.42 \\
        SDHDO~\cite{noh2025sdhdo} & 2D pose map & \underline{81.39} & \underline{0.924} & \underline{16.41} & 7.05 & \underline{23.80} & 43.49 & \underline{64.19} & \underline{0.956} & 6.05 & 6.05 & \underline{24.45} & \underline{9.24} \\
        Ours & 2D pose map & \textbf{49.47} & \textbf{0.948} & \textbf{6.12} & \textbf{4.37} & \textbf{25.86} & \textbf{23.33} & \textbf{38.77} & \textbf{0.970} & \textbf{1.25} & \textbf{3.41} & \textbf{26.93} & \textbf{6.37} \\
        \midrule
        Ours w/o Refinement & 2D pose map & 58.99 & 0.941 & 7.44 & 5.06 & 24.67 & 23.63 & 64.40 & 0.957 & 2.64 & 4.54 & 24.54 & 7.28 \\
        \bottomrule
    \end{tabular}
    %}
    }
    \vspace{-2mm}
    \caption{\textbf{Quantitative comparison for promptable human amodal completion.} We use the same 2D pose map as the user prompt, whereas MCLD uses UV maps following its experimental setting. The best performance for each metric is highlighted in \textbf{bold}, while the second-best is \underline{underlined}. \dag\ indicates injection of the same user prompt via a pre-trained ControlNet for fair comparison. * indicates that values are reported $\times 10^3$ for readability.}
    \vspace{-4mm}
    \label{table1:PHAC}
\end{table*}

%===========================================================
%% Sec 3.4
\subsection{Refinement Network}
\label{sec3.4:refine}

The coarse completion $I_\text{cc}$ generated by $\epsilon_\text{cig}$ exhibits degradation of the visible appearance arising from latent space denoising and subsequent VAE reconstruction. A naïve approach is to construct a baseline composite $I_\text{base}$ by copying pixels from the input image $I_\text{ic}$ in visible regions and from $I_\text{cc}$ in invisible regions, guided by the visible mask $M_\text{v}$:
\begin{equation}
    \label{eq11:i_base}
    I_\text{base} = I_\text{ic} \odot M_\text{v} + I_\text{cc} \odot (1 - M_\text{v}),
\end{equation}
where $\odot$ denotes element-wise multiplication.

While this approach preserves the visible appearance, it often introduces boundary artifacts at mask boundaries and yields an unnatural composite. To address this, we propose an inpainting-based refinement network that keeps the unmasked regions unchanged and performs a few denoising steps in the masked area to reduce boundary artifacts and produce smooth transitions at mask boundaries.

\noindent\textbf{Invisible mask prediction.}
To apply the inpainting-based approach, we require an invisible-region mask $M_\text{iv}$. The masked region specifies where the output should follow the coarse completion $I_\text{cc}$, whereas the unmasked region specifies where information from the incomplete input $I_\text{ic}$ should be preserved. Because a coarse complete image $I_\text{cc}$ is available, we predict the invisible mask using a lightweight U-Net $\mathcal{U}_\text{iv}$. Given the incomplete image $I_\text{ic}$, the coarse completion image $I_\text{cc}$, and the visible mask $M_\text{v}$, U-Net $\mathcal{U}_\text{iv}$ outputs the invisible mask $M_\text{iv}$ as:
\begin{equation}
    \label{eq12:m_iv}
    M_\text{iv} = \mathcal{U}_\text{iv}(I_\text{ic}, I_\text{cc}, M_\text{v}).
\end{equation}
We train the U-Net $\mathcal{U}_\text{iv}$ with a loss function defined as a weighted sum of binary cross-entropy (BCE) and Dice losses~\cite{milletari2016vnet} as follows:
\begin{equation}
    \label{eq13:unet_loss}
    \mathcal{L} = \mathcal{L}_\text{BCE}(M_\text{iv}, M_\text{iv}^*) + \lambda_\text{dice}\mathcal{L}_\text{Dice}(M_\text{iv}, M_\text{iv}^*),
\end{equation}
where $M_\text{iv}^*$ is the GT invisible mask.

Directly applying $M_\text{iv}$ can exclude boundary pixels and introduce boundary artifacts. We therefore use a dilated mask $M_\text{iv}^\prime$ during the refinement process, computed as:
\begin{equation}
    \label{eq14:m_iv'}
    M_\text{iv}^\prime = \text{Dilate}(M_\text{iv}).
\end{equation}

\noindent\textbf{Inpainting-based refinement process.}
Rather than synthesizing entirely new RGB values inside the mask as in conventional inpainting, we aim to preserve the coarse completion and visible appearance while reducing boundary artifacts. To this end, we inject a small amount of noise into the masked region, including the occluded area and its boundary, and run a few denoising steps to refine the completion. With this strategy, we keep the unmasked region unchanged, thereby preserving its appearance, and add only low-magnitude noise to the masked region. This minimizes damage to the coarse completion and effectively removes boundary artifacts. The refinement network is plug-and-play and can be applied to other DM-based methods to improve performance.

The refinement network $\Phi_\text{RF}$ takes as input the baseline composite $I_\text{base}$ from Eq.~\eqref{eq11:i_base} and the dilated invisible mask $M_\text{iv}^\prime$ from Eq.~\eqref{eq14:m_iv'}, and outputs the refined complete image $I_\text{rc}$ as follows:
\begin{equation}
    \label{eq15:i_rc}
     I_\text{rc} = \Phi_\text{RF}(I_\text{base}, M_\text{iv}^\prime, s),
\end{equation}
where $s$ controls the magnitude of noise added to the masked region (noise magnitude $\propto s$). Larger values of $s$ inject more noise, pushing the effective input toward random noise, whereas values of $s$ near zero keep the input close to a clean image.

\section{Experimental Results}

%% Fig 4: qualitative results
\begin{figure*}
    \centering
    \includegraphics[width=0.9\linewidth]{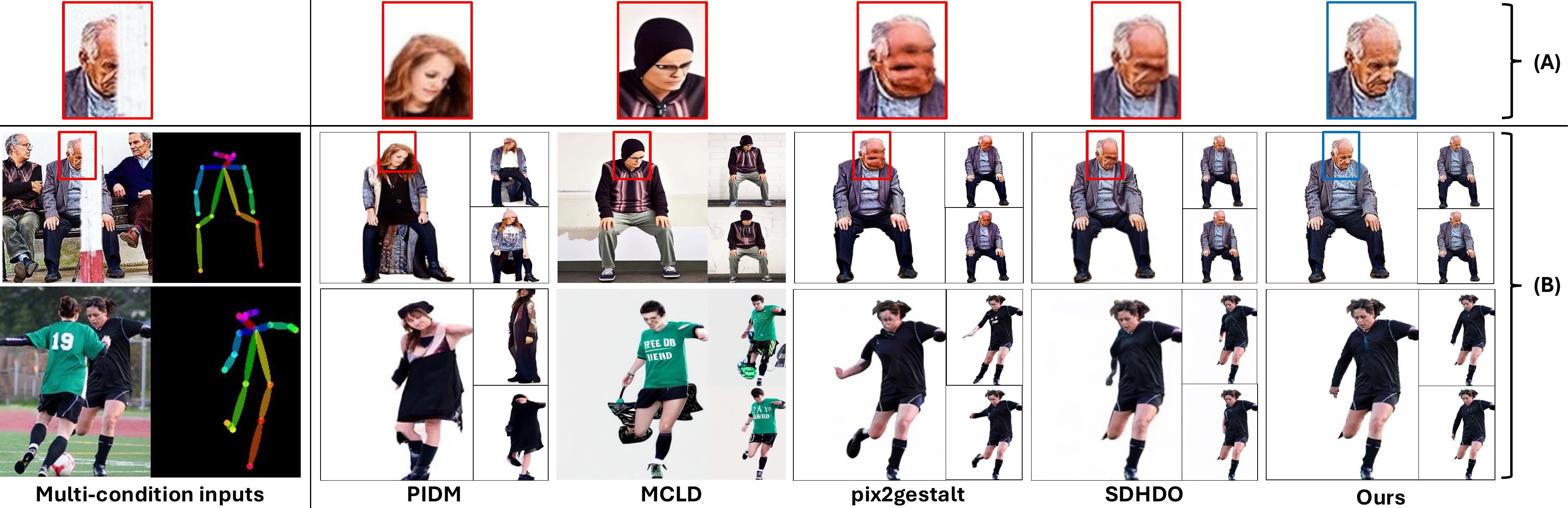}
    \vspace{-2mm}
    \caption{\textbf{Qualitative comparison on the AHP test dataset.} (A) Partial RGB results; (B) Generated images with different seeds. PGPIS baselines (PIDM, MCLD) frequently hallucinate training set appearances. Amodal completion baselines (pix2gestalt, SDHDO) do not preserve the visible appearance and often violate the pose condition. In contrast, our approach yields consistent pose alignment across seeds and preserves the visible regions.}
    \label{fig4:qual}
    \vspace{-4mm}
\end{figure*}

%===========================================================
%% Sec 4.1
\subsection{Datasets}
\label{sec4.1:datasets}

To train the ControlNet, the coarse image generation DM, and the invisible mask prediction U-Net, we constructed a synthetic dataset called OccThuman2.0 based on THuman2.0~\cite{yu2021thuman2}. Specifically, we rendered complete person images from 526 3D human meshes, each from 10 viewpoints. Because the renders contain only upright, centered subjects, we applied flipping, rotation, and scaling augmentations to increase diversity and used the augmented renders as GT complete images. To increase realism, we sampled backgrounds from Places2~\cite{zhou2017places} and composited random object instances from MSCOCO~\cite{lin2014microsoft_coco} in front of the subjects to simulate occlusions, yielding 5,260 occluded person images paired with corresponding GT completions. We sampled occlusion ratios from a Gaussian distribution, as in~\cite{noh2025sdhdo}. We used images from 522 subjects for training and images from the remaining 4 subjects for testing. To assess generalization to in-the-wild images, we also evaluated on the AHP real test set~\cite{zhou2021human_do}, which comprises 56 images with diverse occlusion scenarios. Additional details of the OccThuman2.0 dataset construction are provided in the supplementary material (Sec.~\ref{suppl_sec7.1:occthuman2}).

%===========================================================
%% Sec 4.2
\subsection{Implementation Details}
\label{sec4.2:implement_detail}

We initialize the coarse image generation DM with the weights of the DMs used in prior amodal completion~\cite{ozguroglu2024pix2gestalt} and HAC~\cite{noh2025sdhdo}, and then fine-tune the model. Following~\cite{zhang2023controlnet}, we train the ControlNet by creating a trainable copy of the DM encoder. For ControlNet, when using a single prompt type (pose only or bbox only, $p_\text{po}$, $p_\text{ib}$, $p_\text{eb}$), the prompt input is represented with 3 channels. When using pose and bbox jointly ($p_\text{poib}$, $p_\text{poeb}$), the prompt input contains 6 channels obtained by concatenating the two 3-channel prompts. During training of the DM and ControlNet, we use a constant warm-up schedule~\cite{ho2024sith}. The learning rates are set to $5\times10^{-6}$ for the DM and $5\times10^{-5}$ for ControlNet, with a batch size of 14. The conditioning scale and classifier-free guidance scale are fixed at 1.0 and 2.0, respectively. We apply a dropout rate of 0.05 to the image-conditioning path. In each epoch, we randomly sample one of the 10 views for each subject, yielding 522 images. We train for 1,750 epochs on four RTX A6000 GPUs, with a total training time of approximately 16 hours.

The invisible mask prediction U-Net $\mathcal{U}_{\text{iv}}$ takes as input the coarse completion $I_\text{cc}$ produced by the coarse image generation DM. Following a stochastic inference scheme~\cite{fiche2025mega,ho2024sith}, the DM generates $N=16$ outputs per GT image, and during training $\mathcal{U}_\text{iv}$ randomly samples one of these outputs for supervision. As with the coarse image generation DM, we use 522 images per epoch and train for 40 epochs. We fix the Dice-loss weight at $\lambda_{\text{dice}} = 0.5$. Training runs on a single RTX 3090 and takes approximately 30 minutes.

%===========================================================
%% Sec 4.3
\subsection{Evaluation Metrics}
\label{sec4.3:metrics}

To assess the appearance of the generated complete human images, we evaluate the quality of reconstructed images using Learned Perceptual Image Patch Similarity (LPIPS), Structural Similarity (SSIM), Kernel Inception Distance (KID), Mean Squared Error (MSE), and Peak Signal-to-Noise Ratio (PSNR). Among these, LPIPS, SSIM, and KID evaluate perceptual quality and visual realism, whereas MSE and PSNR focus on pixel-level reconstruction accuracy. For alignment with user prompts, we report a pose metric given by the pixel-wise joint error between poses predicted on the generated images and those predicted on the GT images, which reflects prompt alignment quality.

%===========================================================
%% Sec 4.4
\subsection{Comparison with Existing Methods}
\label{sec4.4:comparison}

We compare our method with the PGPIS baselines PIDM~\cite{bhunia2023pidm} and MCLD~\cite{liu2025mcld}, and the amodal completion methods pix2gestalt~\cite{ozguroglu2024pix2gestalt} and SDHDO~\cite{noh2025sdhdo}. For the PGPIS baselines, we provided user prompts in the format that each model was trained to accept and supplied masked images as input. Because pix2gestalt does not accept user prompts, we additionally evaluate pix2gestalt\dag, which injects the same user prompt via a pre-trained ControlNet for a fair comparison. SDHDO is an HAC method that uses a 2D pose prior. Accordingly, we replace its predicted pose with the user-specified pose used for the other methods.

Table~\ref{table1:PHAC} summarizes quantitative results for the PHAC task. On both the synthetic OccThuman2.0 and the real AHP test set, our method outperforms prior approaches on all metrics. PGPIS methods (PIDM, MCLD), which do not adequately preserve the visible appearance, underperform both the amodal completion baselines (pix2gestalt, SDHDO) and our approach. In particular, despite leveraging a richer UV map prompt, MCLD tends to overfit to the training data and performs poorly on both perceptual and reconstruction metrics. For pix2gestalt, adding a 2D pose map condition yields a small gain on OccThuman2.0 but little change on AHP. In contrast, our method performs well on both synthetic and real data and achieves consistent improvements across all metrics.

Qualitative results in Fig.~\ref{fig4:qual} further demonstrate the effectiveness of our method on PHAC. PIDM and MCLD fail to preserve the visible appearance, instead synthesizing images whose appearance reflects dataset-specific biases learned from DeepFashion~\cite{liu2016deepfashion}. PIDM exhibits severe hallucinations; for example, an older man is transformed into a white woman, and the method frequently fails to satisfy the pose condition. Although MCLD aligns better with the target pose than PIDM, it still produces strong appearance hallucinations. As shown in the bottom row of Fig.~\ref{fig4:qual}(B), an erroneous UV map causes the model to generate a person wearing the occluder’s green clothing instead of the original black garment.

Unlike the PGPIS baselines, pix2gestalt and SDHDO avoid hallucinations in the visible regions and generate plausible person images. However, as in the bottom row of Fig.~\ref{fig4:qual}(B), they often over-generate arms or legs irrespective of the specified pose, producing physically implausible results. As shown in Fig.~\ref{fig4:qual}(A), they also degrade the visible facial appearance and fail to produce meaningful detail, leading to blurry outputs. Across multiple random seeds, their results are pose-inconsistent and frequently misaligned with the pose condition. In contrast, our method successfully reconstructs the visible region and produces high-quality, physically plausible completions for occluded areas, while maintaining consistent alignment with the pose condition across seeds. All results are obtained via random sampling, not cherry-picking. Additional qualitative results are provided in the supplementary material (Sec.~\ref{suppl_sec8.5:add_qual_results}).

%===========================================================
%% Sec 4.5
\subsection{Ablation Experiments}
\label{sec4.5:ablation}

% Table 2: user prompt ablation
\begin{table}[t]
    \centering
    {\scriptsize
    \renewcommand{\arraystretch}{1.2}
    \setlength{\tabcolsep}{3pt}
    \resizebox{\columnwidth}{!}{
    \begin{tabular}{l | cccccc}
        \toprule
        User Prompt
        &
        LPIPS* $\downarrow$ & SSIM $\uparrow$ & MSE* $\downarrow$ & PSNR $\uparrow$ & Joint Err. $\downarrow$ \\
        \midrule
        Pose ($p_\text{po}$) &  49.47 & \textbf{0.948} & \textbf{4.37} & \textbf{25.86} & 23.33 \\  
        Interest bbox ($p_\text{ib}$) & 51.83 & 0.942 & 5.69 & 24.99 & 24.01 \\
        Entire bbox ($p_\text{eb}$) & 52.28 & 0.941 & 5.66 & 25.07 & 28.23 \\
        Pose \& Interest bbox ($p_\text{poib}$) & \textbf{49.35} & \underline{0.947} & \underline{4.56} & \underline{25.69} & \underline{22.15} \\
        Pose \& Entire bbox ($p_\text{poeb}$) & \underline{49.42} & 0.946 & 4.89 & 25.49 & \textbf{21.96} \\
        \bottomrule
    \end{tabular}
        }
    }
    \vspace{-2mm}
    \caption{\textbf{Quantitative comparison across user prompts.} On OccThuman2.0, the pose prompt $p_\text{po}$ yields the best reconstruction metrics, while bbox prompts $(p_\text{ib}, p_\text{eb})$ perform worse and exhibit higher joint errors. Combining pose and bbox $(p_\text{poib}, p_\text{poeb})$ consistently reduces joint error, improving alignment.}
    %\vspace{-2mm}
    \label{table2:user_prompts}
\end{table}

\noindent\textbf{User prompts.}
Table~\ref{table2:user_prompts} presents PHAC results on the OccThuman2.0 dataset with different user prompts. Overall, the pose prompt $p_\text{po}$ provides the most effective single-prompt guidance, whereas bbox-only prompts $(p_\text{ib}, p_\text{eb})$ consistently underperform and exhibit higher joint errors. This is likely because bbox prompts mainly constrain the spatial extent of the synthesis region while leaving body configuration ambiguous, allowing diverse poses within the specified box. Combining pose and bbox $(p_\text{poib}, p_\text{poeb})$ maintains similar perceptual quality to the pose-only setting, while slightly sacrificing reconstruction metrics, but achieves a small and consistent reduction in joint error, indicating improved spatial alignment. Additional analyses are provided in the supplementary material (Sec.~\ref{secs6:ablation_prompt}).

% Table 3: strength ablation
\begin{table}[t]
    \centering
    {\scriptsize
    \renewcommand{\arraystretch}{1.2}
    \setlength{\tabcolsep}{3pt}
    \resizebox{\columnwidth}{!}{
    \begin{tabular}{l | cccc | cccc}
        \toprule
        \raisebox{-2ex}{$s$}
        & \multicolumn{4}{c|}{OccThuman2.0 test dataset}
        & \multicolumn{4}{c}{AHP test dataset} \\
        \cline{2-9}
        & LPIPS* $\downarrow$ & SSIM $\uparrow$ & MSE* $\downarrow$ & PSNR $\uparrow$ & LPIPS* $\downarrow$ & SSIM $\uparrow$ & MSE* $\downarrow$ & PSNR $\uparrow$ \\
        \midrule
        0.1 & \textbf{49.39} & \textbf{0.948} & 4.41 & 25.82 & 39.02 & \textbf{0.970} & 3.46 & 26.84 \\  
        0.3 & 49.49 & \textbf{0.948} & \underline{4.39} & \underline{25.83} & \underline{38.97} & \textbf{0.970} & \underline{3.44} & \underline{26.88} \\
        0.5 & \underline{49.47} & \textbf{0.948} & \textbf{4.36} & \textbf{25.85} & \textbf{38.77} & \textbf{0.970} & \textbf{3.41} & \textbf{26.93} \\
        0.7 & 50.18 & 0.947 & 4.43 & 25.27 & 39.03 & 0.969 & 3.45 & 26.81 \\
        0.9 & 51.91 & 0.943 & 4.77 & 25.39 & 40.02 & 0.967 & 3.68 & 26.33 \\
        \bottomrule
    \end{tabular}
        }
    }
    \vspace{-2mm}
    \caption{\textbf{Effect of the refinement noise strength $s$.} Small $s$ yields limited improvement due to minimal denoising, whereas large $s$ degrades the original appearance.}
    \vspace{-4mm}
    \label{table3:noise_s}
\end{table}

\noindent\textbf{Magnitude of noise.}
Table~\ref{table3:noise_s} reports an ablation of the noise strength $s$ used during refinement, where noise is added inside the invisible region to suppress boundary artifacts. We adopt the standard definition $s\in[0,1]$ (noise magnitude $\propto s$): small $s$ leaves the masked region nearly clean for $\Phi_{\text{RF}}$, whereas large $s$ makes it approach random noise. Insufficient noise fails to resolve boundary artifacts, while excessive noise degrades visible regions and deviates from the coarse appearance. Because refinement has minimal effect on pose alignment, we report only image-quality metrics. On both OccThuman2.0 and AHP, $s=0.5$ yields the best reconstruction and perceptual scores. At $s=0.5$, we run roughly 40\% of the standard denoising steps~\cite{podell2023sdxl}, preserving the visible region and harmonizing the coarse and visible appearances.

% Table 4: refinement ablation
\begin{table}[t]
    \centering
    {\scriptsize
    \renewcommand{\arraystretch}{1.2}
    \setlength{\tabcolsep}{3pt}
    \resizebox{\columnwidth}{!}{
    \begin{tabular}{l | ccccc}
        \toprule
        %\raisebox{-2ex}{User Prompt}
        %& \multicolumn{5}{c}{OccThuman2.0 test dataset} \\
        %\cline{2-6}
        Method
        & LPIPS* $\downarrow$ & KID* $\downarrow$ & MSE* $\downarrow$ & PSNR $\uparrow$ & Joint Err. $\downarrow$ \\
        \midrule
        MCLD~\cite{liu2025mcld} & 59.56 \textcolor{blue}{(49\%)} & 11.77 \textcolor{blue}{(71\%)} & 7.50 \textcolor{blue}{(60\%)} & 23.36 \textcolor{blue}{(27\%)} & 31.41 \textcolor{blue}{(41\%)} \\  
        pix2gestalt~\cite{ozguroglu2024pix2gestalt} & \underline{57.03} \textcolor{blue}{(37\%)} & \underline{8.14} \textcolor{blue}{(51\%)} & \underline{6.21} \textcolor{blue}{(18\%)} & \underline{24.39} \textcolor{blue}{(27\%)} & \underline{30.65} \textcolor{blue}{(16\%)} \\
        SDHDO~\cite{noh2025sdhdo} & 59.41 \textcolor{blue}{(27\%)} & 10.92 \textcolor{blue}{(34\%)} & 6.80 \textcolor{blue}{(4\%)} & 24.12 \textcolor{blue}{(1\%)} & 33.97 \textcolor{blue}{(22\%)} \\
        Ours & \textbf{49.47} \textcolor{blue}{(16\%)} & \textbf{6.12} \textcolor{blue}{(18\%)} & \textbf{4.37} \textcolor{blue}{(14\%)} & \textbf{25.86} \textcolor{blue}{(5\%)} & \textbf{23.33} \textcolor{blue}{(1\%)} \\
        \bottomrule
    \end{tabular}
        }
    }
    \vspace{-2mm}
    \caption{\textbf{Plug-and-play application of the refinement network.} \textcolor{blue}{Blue} marks results improved by refinement, and \% reports the relative improvement over the baseline.}
    \vspace{-2mm}
    \label{table4:plug&play}
\end{table}

\noindent\textbf{Refinement network.}
With an appropriate choice of $s$, the proposed refinement network can be applied in a plug-and-play manner to improve the outputs of other generative methods. Table~\ref{table4:plug&play} presents results after applying our refinement network to existing PGPIS and amodal completion baselines. The refinement improves not only perceptual quality and reconstruction metrics but also pose-related measures. In particular, for MCLD~\cite{liu2025mcld}, which exhibited larger errors, our refinement reduces MSE and KID by approximately 60--70\%. Baseline (pre-refinement) scores for the existing methods are reported in Table~\ref{table1:PHAC}, and the last row isolates the effect of our refinement network on the proposed method. Across both datasets, refinement improves all metrics. Moreover, even before refinement, our method already outperforms prior methods.

\section{Conclusion}

We introduced a new task, \emph{promptable human amodal completion (PHAC)}, which completes occluded human images while satisfying both visible appearance constraints and user-specified prompts. Our framework injects point-based pose and bbox prompts via ControlNet modules and fine-tunes only the cross-attention blocks, achieving strong prompt alignment. Additionally, we propose an inpainting-based refinement module that leaves the visible regions unchanged, applies a few denoising steps to the completed areas, and maintains boundary continuity. Because it operates on generic inputs, the module can be used as a plug-and-play component with existing diffusion-based completion methods. Experiments on amodal completion and PGPIS benchmarks show that our approach achieves better prompt alignment and produces more physically plausible, higher-quality completions than prior work.

% \vspace{0.1cm}
\vspace{0.8cm}
\noindent\textbf{Acknowledgement.} This work was supported by Institute of Information \& Communications Technology Planning \& Evaluation (IITP) grant funded by the Korea government (MSIT) (No. RS-2023-00219700, Development of FACS-compatible Facial Expression Style Transfer Technology for Digital Human).

\renewcommand{\thesection}{S\arabic{section}}
\renewcommand{\thefigure}{S\arabic{figure}}
\renewcommand{\thetable}{S\arabic{table}}
\renewcommand{\theequation}{S\arabic{equation}}

\clearpage
\maketitlesupplementary

%% Fig 5: user prompts
\begin{strip}
     %\vspace{-2.5em}
     \centering
     \includegraphics[width=0.95\linewidth]{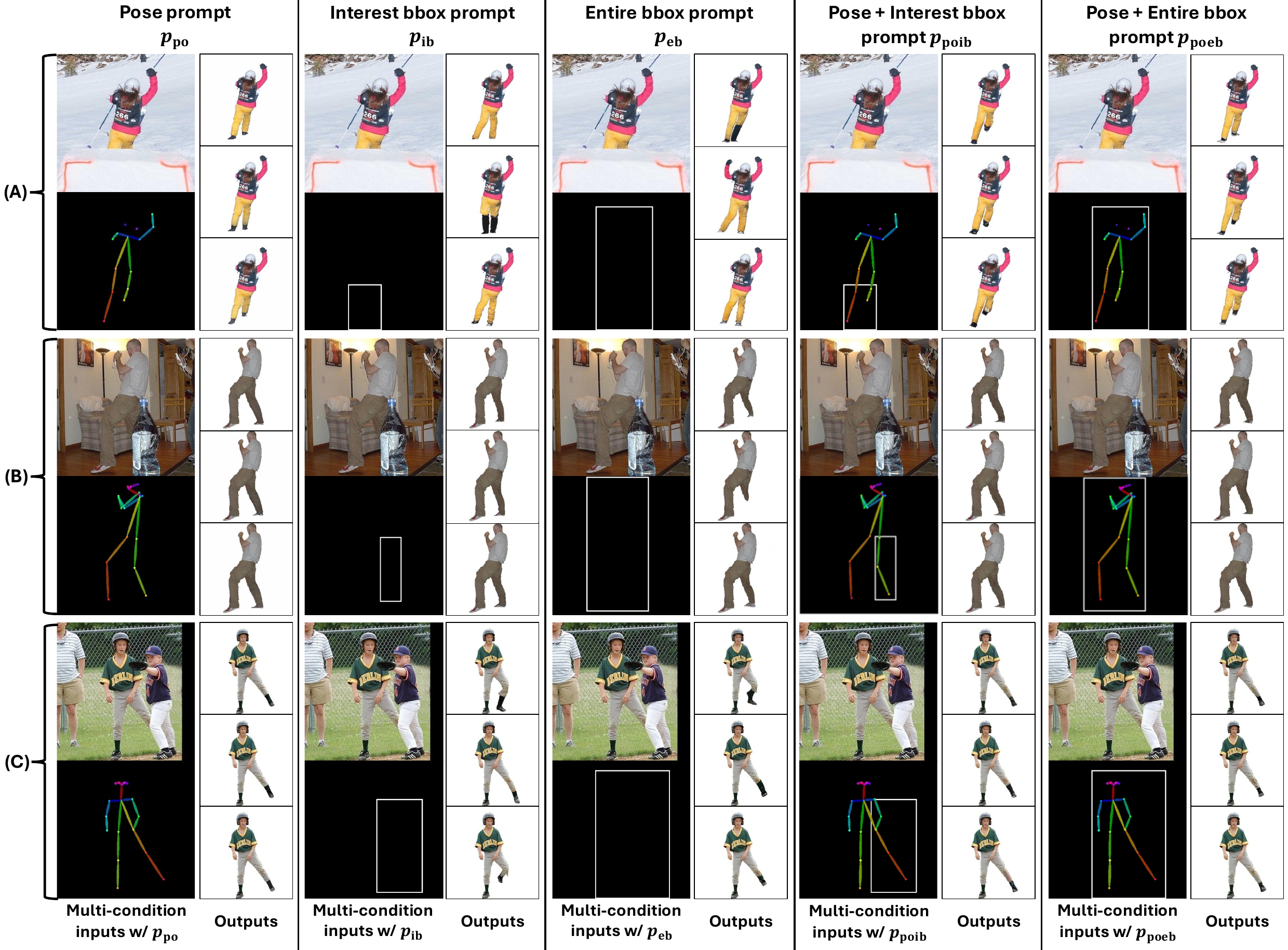}
     \vspace{-3mm}
     \captionof{figure}{\textbf{Qualitative results with different user prompts.} Examples (A, B, C) are from the AHP real test dataset. For each prompt type, three samples are generated using the same random seeds for a fair comparison. In (A), the three samples in the $p_\text{po}$ column are generated using seeds 42, 616, and 2026, and the same set of seeds is reused for the remaining prompt types ($p_\text{ib}$, $p_\text{eb}$, $p_\text{poib}$, $p_\text{poeb}$), ensuring that images within the same row are generated with identical seeds.}
     \label{suppl_fig5:user_prompt}
     %\vspace{-2mm}
\end{strip}

% Table 5: user prompts
\begin{table*}[t]
    \centering
    {\scriptsize
    \renewcommand{\arraystretch}{1.2}
    \setlength{\tabcolsep}{3pt}
    \begin{tabular}{l|cccccc|cccccc}
        \toprule
        \raisebox{-2ex}{User Prompt}
        & \multicolumn{6}{c|}{OccThuman2.0 test dataset (synthetic)}
        & \multicolumn{6}{c}{AHP test dataset (real)} \\
        \cline{2-7} \cline{8-13}
        & LPIPS* $\downarrow$ & SSIM $\uparrow$ & KID* $\downarrow$ & MSE* $\downarrow$ & PSNR $\uparrow$ & Joint Err. $\downarrow$
        & LPIPS* $\downarrow$ & SSIM $\uparrow$ & KID* $\downarrow$ & MSE* $\downarrow$ & PSNR $\uparrow$ & Joint Err. $\downarrow$ \\
        \midrule
        Pose ($p_\text{po}$) & 49.47 & \textbf{0.948} & \underline{6.12} & \textbf{4.37} & \textbf{25.86} & 23.33 & 38.77 & \underline{0.970} & \textbf{1.25} & \textbf{3.41} & \textbf{26.93} & 6.37 \\
        Interest bbox ($p_\text{ib}$) & 51.83 & 0.942 & 6.89 & 5.69 & 24.99 & 24.01 & 41.03 & 0.964 & 1.44 & 4.87 & 25.99 & 11.18 \\
        Entire bbox ($p_\text{eb}$) & 52.28 & 0.941 & 6.32 & 5.66 & 25.07 & 28.23 & 41.35 & 0.963 & 1.37 & 5.04 & 25.78 & 11.84 \\
        Pose \& Interest bbox ($p_\text{poib}$) & \textbf{49.35} & \underline{0.947} & \textbf{6.03} & \underline{4.56} & \underline{25.69} & \underline{22.15} & \underline{38.67} & \underline{0.970} & \underline{1.31} & \underline{3.42} & 26.60 & \underline{6.21} \\
        Pose \& Entire bbox ($p_\text{poeb}$) & \underline{49.42} & 0.946 & \underline{6.12} & 4.89 & 25.49 & \textbf{21.96} & \textbf{38.51} & \textbf{0.971} & 1.33 & 3.48 & \underline{26.74} & \textbf{6.15} \\
        \bottomrule
    \end{tabular}
    }
    \vspace{-2mm}
    \caption{\textbf{Quantitative comparison across different user prompts.} }
    \vspace{-2mm}
    \label{suppl_table5:user_prompt}
\end{table*}

%===========================================================
\section{Additional Analysis of User Prompts}
\label{secs6:ablation_prompt}

Table~\ref{table2:user_prompts} in the main paper reports the user prompt ablation results on OccThuman2.0. In this section, we extend the evaluation by presenting additional quantitative and qualitative results on the AHP real test dataset, including the KID metric. Based on these results, we analyze each prompt in terms of performance and user effort by reporting the marginal gain in joint error per additional input point, enabling a normalized comparison across prompts.

%===========================================================
%% Sec 6.1 
\subsection{Additional Qualitative Analysis}
\label{suppl_sec6.1:prompt_qual}

Qualitative results for the same input image under different user prompts are shown in Fig.~\ref{suppl_fig5:user_prompt}. While the pose prompt $p_\text{po}$ provides joint-level pose constraints, it does not directly constrain the synthesis region. As shown in the $p_\text{po}$ column of Fig.~\ref{suppl_fig5:user_prompt}(A), the occluded right leg generally follows the 2D pose map, but its reconstructed length varies across samples.

The interest-region bbox prompt $p_\text{ib}$ provides a direct constraint on the synthesis region but offers less control over pose compared with the pose prompt $p_\text{po}$. In the $p_\text{ib}$ column of Fig.~\ref{suppl_fig5:user_prompt}(A, C), the occluded leg is synthesized within the bbox, and the resulting poses show diverse variations within this constraint. The entire-region bbox prompt $p_\text{eb}$ provides an even weaker constraint on the occluded regions than the interest-region bbox $p_\text{ib}$. Consequently, the model sometimes generates a left arm within the entire-region bbox (the $p_\text{eb}$ column of Fig.~\ref{suppl_fig5:user_prompt}(A)), and when the input already satisfies this bbox constraint, it may fail to reconstruct the occluded leg (the $p_\text{eb}$ column of Fig.~\ref{suppl_fig5:user_prompt}(B)). Similar to the interest-region bbox $p_\text{ib}$, the poses remain variable within the bbox, as seen in the $p_\text{eb}$ column of Fig.~\ref{suppl_fig5:user_prompt}(C).

This issue can be mitigated by providing both a bbox prompt and a pose prompt as the user prompt. Regardless of bbox type, combining a pose with either an interest-region or entire-region bbox $(p_\text{poib}, p_\text{poeb})$ yields images that are consistent across samples and satisfy both the pose and bbox constraints simultaneously.

%===========================================================
%% Sec 6.2
\subsection{Additional Quantitative Results}
\label{suppl_sec6.2:prompt_quan}

For a more detailed analysis, we evaluate our method on both the OccThuman2.0 and AHP test datasets and report the results for all metrics described in Sec.~\ref{sec4.3:metrics} of the main paper in Table~\ref{suppl_table5:user_prompt}.

Across both datasets, bbox prompts $(p_\text{ib}, p_\text{eb})$ yield higher joint error than the pose prompt $p_\text{po}$. This pattern is consistent with the qualitative examples in Fig.~\ref{suppl_fig5:user_prompt}, where bbox prompting allows diverse poses within the specified region. In addition, the entire-region bbox prompt $p_\text{eb}$ generally performs worse than the interest-region bbox prompt $p_\text{ib}$, as shown in Fig.~\ref{suppl_fig5:user_prompt}, where the less specific entire-region constraint can lead to incomplete reconstructions of occluded parts or spurious content within the bbox.

Combining pose and bbox prompts $(p_\text{poib}, p_\text{poeb})$ yields overall performance comparable to the pose-only setting. In addition, it provides a small but consistent reduction in joint error across both datasets, regardless of bbox type.

\begin{table}[t]
  \centering
  {\scriptsize
  \renewcommand{\arraystretch}{1.2}
  \begin{tabular}{l|ccccc}
    \toprule
    & $p_{\text{po}}$ & $p_{\text{ib}}$ & $p_{\text{eb}}$ & $p_{\text{poib}}$ & $p_{\text{poeb}}$ \\
    \midrule
    $\Delta$ JE & 6.43 & 5.75 & 1.53 & \underline{7.61} & \textbf{7.80} \\
    $\Delta$ JE pp & \underline{1.07} & \textbf{2.86} & 0.77 & 0.95 & 0.98 \\
    \bottomrule
  \end{tabular}
  }
  \vspace{-2mm}  
  \caption{\textbf{Quantitative comparison of prompt efficiency.} We report the absolute gain in joint error ($\Delta$JE) and the per-point gain ($\Delta$JE pp) for different user prompts. While pose-based prompts yield larger absolute gains, the interest-region bbox prompt $p_{\text{ib}}$ achieves the largest per-point gain, indicating efficient improvements with minimal user input.}
  \vspace{-2mm}
  \label{suppl_table6:prompt_eff}
\end{table}

%===========================================================
%% Sec 6.3
\subsection{Prompt Efficiency under User Effort}
\label{suppl_sec6.3:prompt_eff}

Different prompts provide distinct levels of structural guidance and require different amounts of user input. The pose prompt $p_{\text{po}}$ provides joint-level structural cues, but under severe occlusion it may require many input points to specify a reliable pose. In contrast, bbox prompts $(p_{\text{ib}}, p_{\text{eb}})$ require only two input points, providing a lightweight way to localize the target synthesis region.

To quantify prompt efficiency with respect to user effort, we report both the absolute gain in joint error ($\Delta$JE) and the gain per input point ($\Delta$JE pp) in Table~\ref{suppl_table6:prompt_eff}. Pose-based prompts $(p_{\text{po}}, p_{\text{poib}}, p_{\text{poeb}})$ yield larger absolute gains, reflecting the benefit of stronger structural constraints. The interest-region bbox prompt $p_{\text{ib}}$ achieves the highest per-point gain, indicating efficient improvements with minimal user input. In contrast, the entire-region bbox prompt $p_{\text{eb}}$ provides the weakest constraint, as discussed in Secs.~\ref{suppl_sec6.1:prompt_qual} and~\ref{suppl_sec6.2:prompt_quan}. This encourages diverse plausible generations within the bbox; consequently, $p_{\text{eb}}$ yields the lowest gains in both $\Delta$JE and $\Delta$JE pp when evaluated against the ground-truth pose. Finally, combining pose and bbox prompts $(p_{\text{poib}}, p_{\text{poeb}})$ results in higher absolute gains than single-prompt settings, but lower per-point gains than $p_{\text{po}}$ and $p_{\text{ib}}$, reflecting the increased user effort required by the combined prompts.

Overall, supporting multiple prompt types enables flexible trade-offs between performance and user effort, allowing users to select prompts that best match the difficulty of occlusion and the desired level of interaction.

%% Fig 6: dataset pipeline
\begin{figure}
    \centering
    \includegraphics[width=1.0\linewidth]{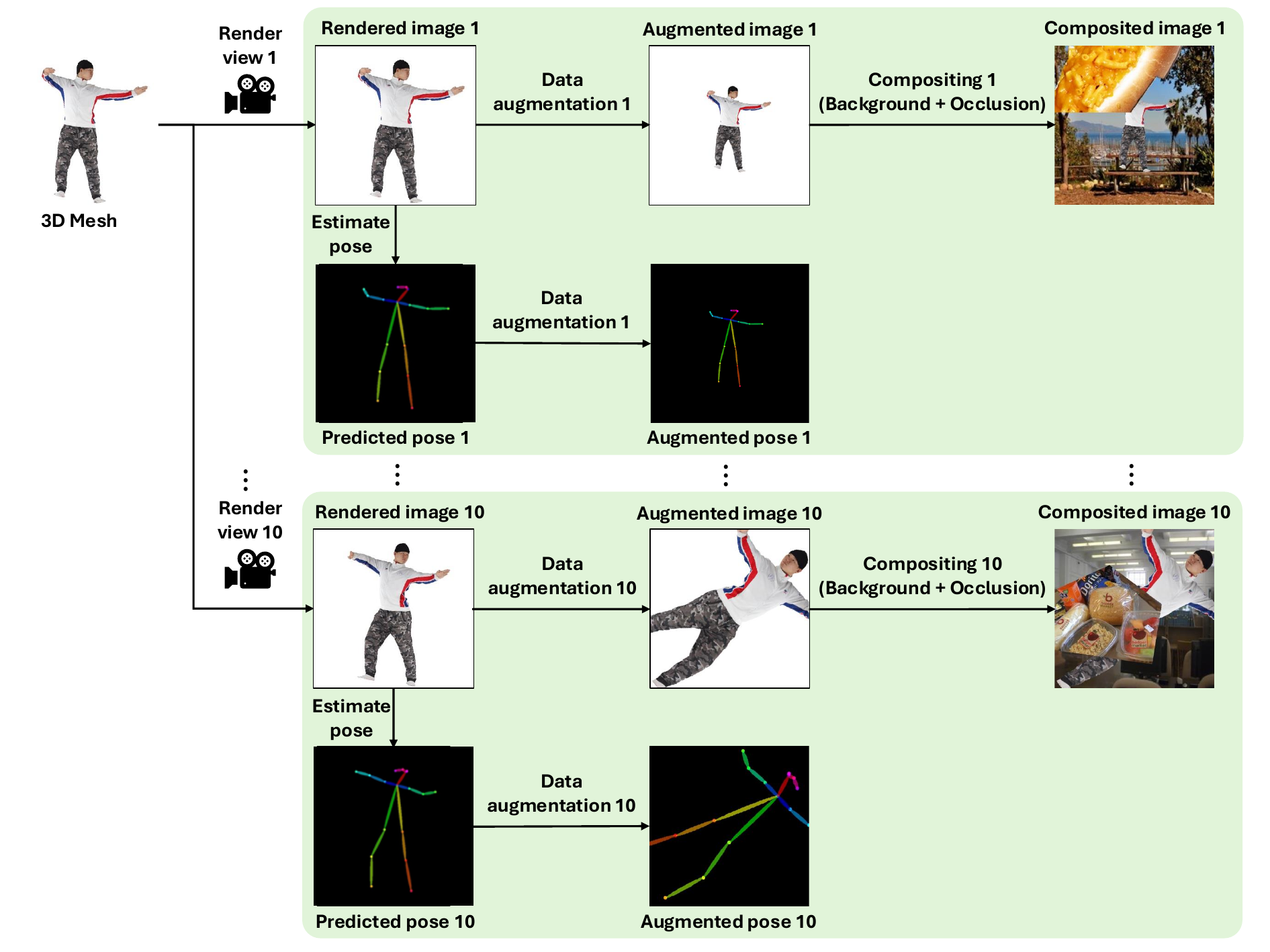}
    \vspace{-6mm}
    \caption{\textbf{Construction pipeline of the OccThuman2.0 dataset.} For each THuman2.0 mesh, we render 10 views and apply per-view data augmentations and background-occlusion compositing to generate 10 composited images. Repeating this process across all 526 meshes produces a total of 5,260 composited images.}
    \label{suppl_fig6:occthuman2}
    \vspace{-2mm}
\end{figure}

%===========================================================
\section{Detailed Implementations}

%===========================================================
%% Sec 7.1 
\subsection{OccThuman2.0 Dataset}
\label{suppl_sec7.1:occthuman2}

As discussed in Sec.~\ref{sec4.1:datasets} of the main paper, we construct the OccThuman2.0 dataset from THuman2.0~\cite{yu2021thuman2} for training and evaluation. Fig.~\ref{suppl_fig6:occthuman2} illustrates the dataset construction pipeline for 10 views of a single subject. For each subject, we first render 10 views from the corresponding 3D mesh, following prior work on clothed human reconstruction~\cite{ho2024sith,xiu2022icon,hong2024chr_smv}. From the rendered images, we estimate 2D human poses using OpenPose~\cite{cao2019openpose}. We then apply the same augmentations, including horizontal flip, vertical flip, rotation, and scaling, to both the images and the poses. Each augmentation is applied independently with a probability of 30\%. When rotation is applied, we sample a single angle $\theta'$ from a uniform distribution $\theta' \sim \mathcal{U}[0^\circ, 360^\circ)$. When scaling is applied, we sample a single scale factor $s' \sim \mathcal{U}[0.5, 1.5]$ to prevent the person from becoming excessively small or large. At this stage, the rendered images cover diverse person orientations and scales but contain no backgrounds or occlusions.

We composite backgrounds using images randomly sampled from Places2~\cite{zhou2017places} and synthesize occlusions by overlaying randomly sampled MSCOCO~\cite{lin2014microsoft_coco} object instances onto the images, following previous work on amodal completion~\cite{zhan2020sssd,ozguroglu2024pix2gestalt,noh2025sdhdo}. Since the last four subjects in THuman2.0 (0522-0525) are scans of the same person, we use the corresponding 40 images (4 subjects $\times$ 10 views) for testing and the remaining 5,220 images corresponding to 522 subjects (0000-0521) for training. For each subject, we sample a target occlusion ratio from a Gaussian distribution and ensure that the final set of ratios follows this target distribution. We use the GT complete images (before background and occlusion compositing) to train the coarse image generation diffusion model $\epsilon_\text{cig}$ and the ControlNet $\Phi_\text{CN}$, and the GT invisible masks to train the invisible mask prediction U-Net $\mathcal{U}_\text{iv}$.

%===========================================================
%% Sec 7.2
\subsection{Invisible Mask Prediction U-Net}
\label{suppl_sec7.2:u_iv}

To train the invisible mask prediction U-Net $\mathcal{U}_\text{iv}$, we first train the coarse image generation DM and the ControlNet. Following Sec.~\ref{sec4.2:implement_detail} of the main paper, we generate 16 coarse complete images $I_\text{cc}$ for each training input image $I_\text{ic}$. During the training of $\mathcal{U}_\text{iv}$, we randomly select one of these 16 coarse complete images as input. We use the AdamW optimizer with a learning rate of $1 \times 10^{-4}$ and a weight decay of $1 \times 10^{-4}$. A single invisible mask prediction step takes approximately 0.02 seconds per image (about 50 fps) on a single RTX 3090 GPU. During training, each term of the loss in Eq.~\ref{eq13:unet_loss} of the main paper is computed as follows:
\begin{equation}
    \label{eq16:bce_loss}
    \mathcal{L}_\text{BCE} = -\frac{1}{N}\sum_{i=1}^N \big[p^*_i \log p_i + (1 - p^*_i)\log(1 - p_i)\big],
\end{equation}
\begin{equation}
   \label{eq17:dice_loss}
   \mathcal{L}_\text{Dice} = 1 - \frac{2\sum_{i=1}^N p_i p^*_i + \delta}{\sum_{i=1}^N p_i + \sum_{i=1}^N p^*_i + \delta},
\end{equation}
where $p_i = M_\text{iv}(i)$ and $p^*_i = M_\text{iv}^*(i)$ denote the predicted and GT invisible mask values at pixel $i$, respectively, and $N$ denotes the number of pixels. $\delta = 1 \times 10^{-6}$ is a small positive constant added for numerical stability to prevent division by zero.

%===========================================================
%% Sec 7.3
\subsection{Refinement Network}
\label{suppl_sec7.3:phi_rf}

We use a pre-trained Stable Diffusion XL (SDXL) model~\cite{podell2023sdxl} as the refinement network $\Phi_\text{RF}$. Since SDXL is optimized for a resolution of $1024 \times 1024$, we resize the baseline composite image $I_\text{base}$ to $1024 \times 1024$ before feeding it into $\Phi_\text{RF}$. For evaluation, all generated images are resized to a fixed resolution to ensure a fair comparison across methods. We fix the number of refinement steps to 20 and set the noise strength to $s = 0.5$, based on the ablation study in Table~\ref{table3:noise_s} of the main paper. With these settings, denoising begins at timestep $t_0 = 380$. We set the classifier-free guidance scale to 1.5 and use the \nolinkurl{diffusers/stable-diffusion-xl-1.0-inpainting-0.1} checkpoint from the Diffusers library~\cite{von-platen-etal-2022-diffusers}. The refinement step takes approximately 4 seconds per image (about 0.25 fps) on a single RTX 3090 GPU.

%% Fig 7: MCLD inference
\begin{figure}
    \centering
    \includegraphics[width=\linewidth]{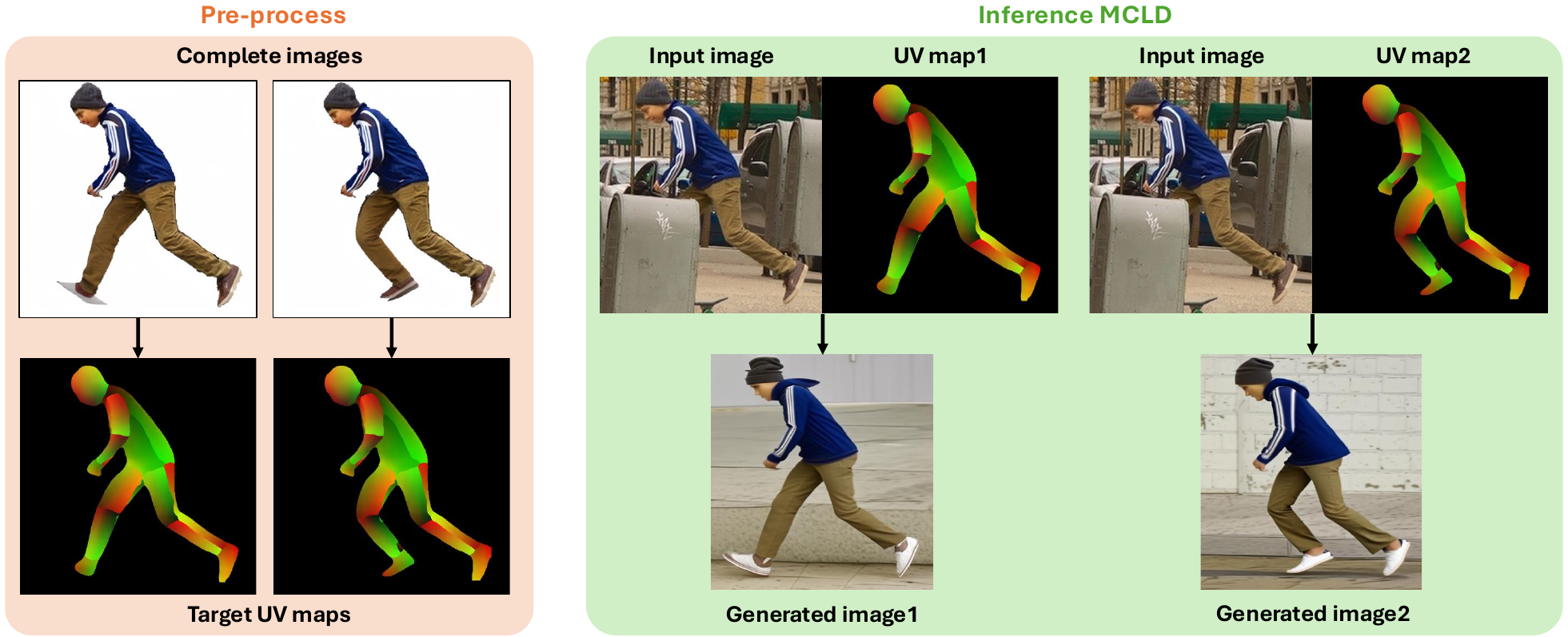}
    \vspace{-6mm}
    \caption{\textbf{Prompt image for MCLD.} MCLD uses a DensePose-based UV map as its conditioning input instead of a 2D pose map. We obtain the target UV map from the occlusion-free complete image and perform MCLD inference conditioned on this UV map.}
    \label{suppl_fig7:mcld}
    \vspace{-2mm}
\end{figure}

%===========================================================
%% Sec 7.4
\subsection{Prompt Image}
\label{suppl_sec7.4:prompt_image}

As described in Sec.~\ref{sec3.3:CIG} of the main paper, our approach takes as input simple, user-friendly prompts $P$ consisting of only a few points. For the pose prompt $p_\text{po}$, the input consists of the indices and 2D coordinates of additional joints beyond the visible ones. Given these user-specified joints, we convert the 2D pose from the OpenPose keypoint format to the COCO format~\cite{lin2014microsoft_coco} and, following ControlNet~\cite{zhang2023controlnet}, render it as a 2D pose map with a fixed color mapping and parent-child relations. This pose map is used as the prompt image $I_\text{p}$.

For the bbox prompts $p_\text{ib}$ and $p_\text{eb}$, the input consists of two points corresponding to the top-left and bottom-right corners. We construct the corresponding axis-aligned bbox and visualize its boundary with a fixed thickness, setting boundary pixels to 255 (white) and all other pixels to 0 (black). When the thickness is set to 1, the resulting bbox image has a one-pixel boundary. In practice, we fix the thickness to 4 for both training and inference. To improve robustness to noisy user-provided bbox prompts, we augment the bbox during training by randomly scaling it and jittering its corner coordinates with 50\% probability. When using both pose and bbox prompts $(p_\text{poib}, p_\text{poeb})$, we concatenate their corresponding prompt images along the channel dimension to form the final prompt image $I_\text{p}$.

As noted in Sec.~\ref{suppl_sec6.1:prompt_qual}, because most previous methods accept a 2D pose map as input, we use only the prompt image derived from the pose prompt $p_\text{po}$ for comparison. Because MCLD~\cite{liu2025mcld} is conditioned on a UV map~\cite{guler2018densepose}, we use the UV map extracted from the occlusion-free complete image as the prompt image $I_\text{p}$. Fig.~\ref{suppl_fig7:mcld} shows the UV map prompt used to generate the MCLD results in Fig.~\ref{fig1:teaser} of the main paper. The UV map provides not only pose information but also regional coverage, supplying cues that help the model generate a plausible human shape. Such regional information is not available in a 2D pose map.

%===========================================================
\section{Additional Results}

% Table 7: invisible mask prediction
\begin{table}[t]
    \centering
    {\scriptsize
    \renewcommand{\arraystretch}{1.2}
    \setlength{\tabcolsep}{3pt}
    \begin{tabular}{l|cc|cc}
        \toprule
        \raisebox{-2ex}{Method}
        & \multicolumn{2}{c|}{OccThuman2.0}
        & \multicolumn{2}{c}{AHP} \\
        \cline{2-5}
        & mIoU $\uparrow$ & L1* $\downarrow$
        & mIoU $\uparrow$ & L1* $\downarrow$ \\
        \midrule
        pix2gestalt~\cite{ozguroglu2024pix2gestalt} & 48.26 & 39.74 & 48.97 & 22.04 \\
        SDHDO~\cite{noh2025sdhdo} & 51.86 & 33.58 & 51.47 & 21.35 \\
        \midrule
        Ours & 87.84 & 6.04 & 77.87 & 7.89 \\
        \bottomrule
    \end{tabular}
    }
    \vspace{-2mm}
    \caption{\textbf{Quantitative results for invisible mask prediction.} Previous amodal completion methods~\cite{ozguroglu2024pix2gestalt,noh2025sdhdo} do not use the coarse complete image $I_\text{cc}$ as input, so their results are not directly comparable.}
    \vspace{-2mm}
    \label{suppl_table7:invis_mask}
\end{table}

%% Fig 8: invisible mask prediction
\begin{figure}
    \centering
    \includegraphics[width=1.0\linewidth]{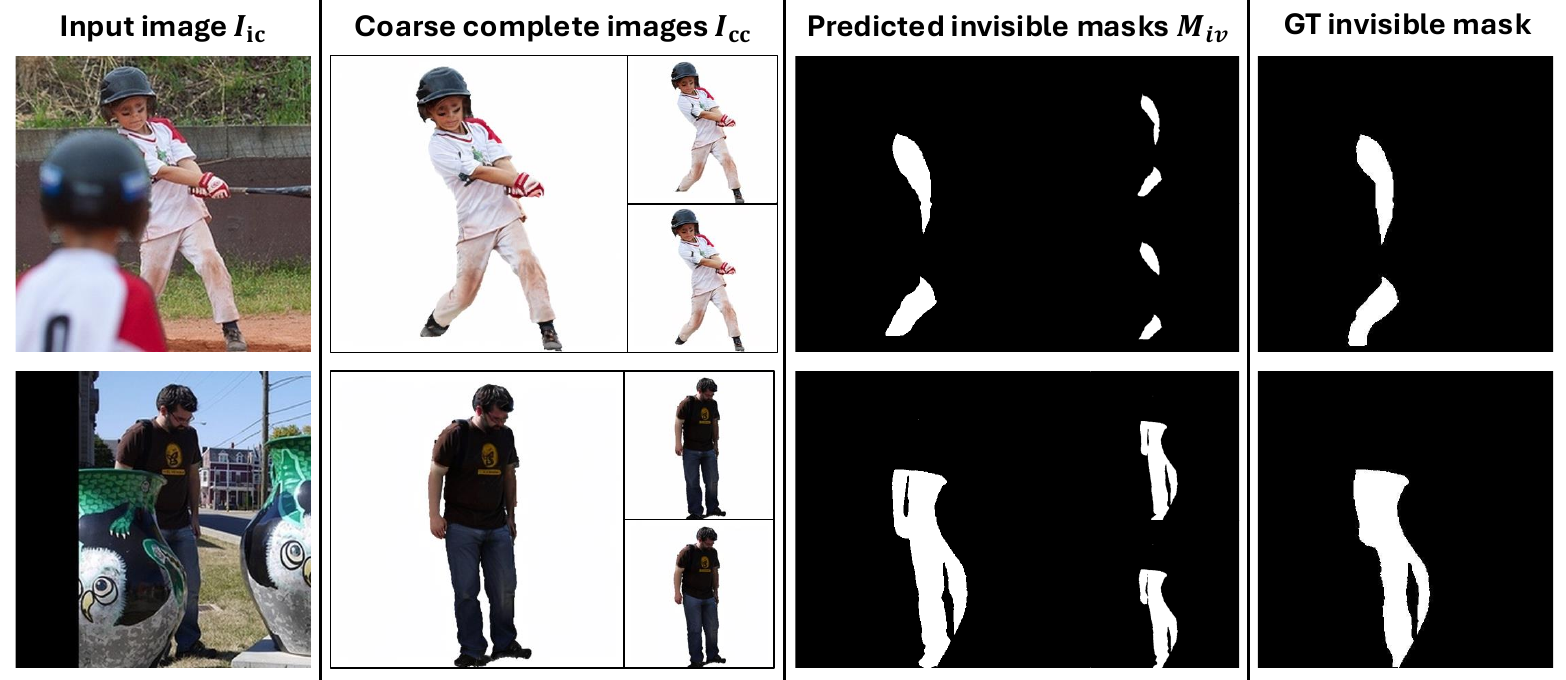}
    \vspace{-6mm}
    \caption{\textbf{Qualitative results for invisible mask prediction.} Given the coarse complete image $I_\text{cc}$, $\mathcal{U}_\text{iv}$ predicts invisible masks $M_\text{iv}$ that closely match the ground truth.}
    \label{suppl_fig8:invis_mask}
    \vspace{-2mm}
\end{figure}

%===========================================================
%% Sec 8.1
\subsection{Invisible Mask}
\label{suppl_sec8.1:invisible_mask}

In this section, we present quantitative and qualitative results for the invisible mask prediction U-Net $\mathcal{U}_\text{iv}$. To evaluate performance, we report invisible mask prediction results on both OccThuman2.0 and the AHP real test dataset, measuring performance using mean Intersection over Union (mIoU) and L1 loss. Table~\ref{suppl_table7:invis_mask} compares our method with previous amodal completion methods~\cite{ozguroglu2024pix2gestalt,noh2025sdhdo}. These methods do not take the coarse complete image $I_\text{cc}$ as input; therefore, their results are not directly comparable and are provided for reference only. Predicting the invisible mask $M_\text{iv}$ from the incomplete image $I_\text{ic}$ and the visible mask $M_\text{v}$ alone is under-constrained, whereas coarse completion provides $I_\text{cc}$ as a hypothesis over the invisible regions. Conditioning $M_\text{iv}$ on $I_\text{cc}$ therefore encourages consistency between coarse completion and invisible-region localization.

As shown in Fig.~\ref{suppl_fig8:invis_mask}, when using the coarse complete image $I_\text{cc}$ as input, $\mathcal{U}_\text{iv}$ predicts invisible masks that closely match the GT invisible mask. To improve robustness to noisy or structurally imperfect coarse completions, we train $\mathcal{U}_\text{iv}$ with multiple randomly sampled $I_\text{cc}$ candidates per GT mask. For a single input image $I_\text{ic}$, the stochastic sampling scheme generates multiple distinct coarse complete images $I_\text{cc}$, and we predict an invisible mask $M_\text{iv}$ for each sample. Since the input image is identical across samples, the GT invisible mask is the same for all of them.

%% Fig 9: refinement process
\begin{figure}
    \centering
    \includegraphics[width=1.0\linewidth]{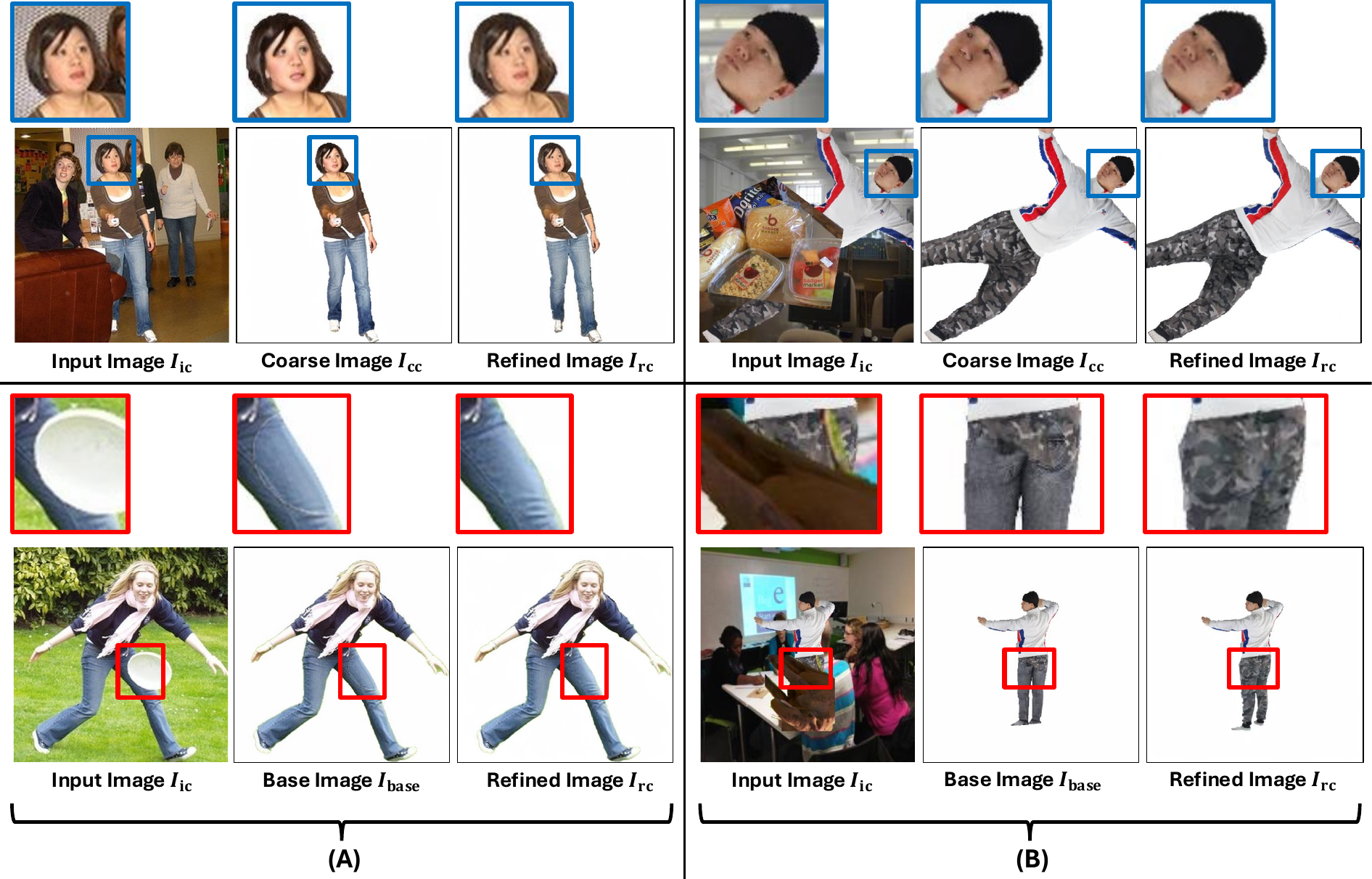}
    \vspace{-6mm}
    \caption{\textbf{Qualitative results for the refinement process.} (A) Results on the AHP dataset and (B) results on the OccThuman2.0 dataset. The \textcolor{blue}{blue bboxes} compare the facial regions of the input image $I_\text{ic}$, the coarse complete image $I_\text{cc}$, and the refined image $I_\text{rc}$ to assess preservation of visible appearance. The \textcolor{red}{red bboxes} compare the boundary regions of the input image $I_\text{ic}$, the base composite image $I_\text{base}$, and the refined image $I_\text{rc}$ to verify smooth blending without boundary artifacts.}
    \label{suppl_fig9:refinement}
    \vspace{-2mm}
\end{figure}

% Table 8: zero prompt
\begin{table*}[t]
    \centering
    {\scriptsize
    \renewcommand{\arraystretch}{1.2}
    \setlength{\tabcolsep}{3pt}
    \begin{tabular}{l|c|cccccc|cccccc}
        \toprule
        \raisebox{-2ex}{Method} & \raisebox{-2ex}{User Prompt}
        & \multicolumn{6}{c|}{OccThuman2.0 test dataset (synthetic)}
        & \multicolumn{6}{c}{AHP test dataset (real)} \\
        \cline{3-8} \cline{9-14}
        &
        & LPIPS* $\downarrow$ & SSIM $\uparrow$ & KID* $\downarrow$ & MSE* $\downarrow$ & PSNR $\uparrow$ & Joint Err. $\downarrow$
        & LPIPS* $\downarrow$ & SSIM $\uparrow$ & KID* $\downarrow$ & MSE* $\downarrow$ & PSNR $\uparrow$ & Joint Err. $\downarrow$ \\
        \midrule
        pix2gestalt~\cite{ozguroglu2024pix2gestalt} & - & 90.11 & 0.911 & 16.51 & 7.58 & 22.63 & 36.65 & 75.73 & 0.942 & 5.98 & 5.22 & 24.06 & 10.96 \\
        pix2gestalt~\cite{ozguroglu2024pix2gestalt}\dag & 2D pose map & 88.58 & 0.914 & 16.75 & 6.93 & 22.94 & 31.37 & 75.25 & 0.943 & 6.35 & 4.87 & 24.25 & 10.42 \\
        SDHDO~\cite{noh2025sdhdo} & 2D pose map & 81.39 & 0.924 & 16.41 & 7.05 & 23.80 & 43.49 & 64.19 & 0.956 & 6.05 & 6.05 & 24.45 & \underline{9.24} \\
        Ours & - & \underline{53.19} & \underline{0.939} & \underline{6.23} & \underline{5.69} & \underline{25.03} & \underline{29.76} & \underline{41.99} & \underline{0.964} & \underline{1.60} & \underline{4.79} & \underline{25.86} & 13.23 \\
        Ours & 2D pose map & \textbf{49.47} & \textbf{0.948} & \textbf{6.12} & \textbf{4.37} & \textbf{25.86} & \textbf{23.33} & \textbf{38.77} & \textbf{0.970} & \textbf{1.25} & \textbf{3.41} & \textbf{26.93} & \textbf{6.37} \\
        \bottomrule
    \end{tabular}
    }
    \vspace{-2mm}
    \caption{\textbf{Quantitative comparison for PHAC.} This table largely mirrors Table~\ref{table1:PHAC} of the main paper, but additionally includes the results of our method evaluated without a user prompt. To avoid redundancy, PGPIS results are omitted. \dag\ indicates the use of the same user prompt injected through a pre-trained ControlNet.}
    \vspace{-2mm}
    \label{suppl_table8:zero_prompt}
\end{table*}

%===========================================================
%% Sec 8.2
\subsection{Refinement Process}
\label{suppl_sec8.2:refinement}

Our refinement process addresses two issues: degradation of the visible appearance in the coarse complete image $I_\text{cc}$ and boundary artifacts introduced by naïve compositing to form the base composite image $I_\text{base}$. The degradation in the visible region can be partially mitigated by compositing the coarse complete image $I_\text{cc}$ with the input image $I_\text{ic}$ using the visible region mask $M_\text{v}$ to obtain the base composite image $I_\text{base}$, as in Eq.~\ref{eq11:i_base} of the main paper. However, the base composite image $I_\text{base}$ still suffers from boundary artifacts and fails to blend the appearance smoothly across regions. To address these issues, we adopt an inpainting-based refinement process.

As shown in the blue bboxes in Fig.~\ref{suppl_fig9:refinement}, the coarse complete image $I_\text{cc}$ exhibits degradation in fine-grained facial details. In contrast, the refined complete image $I_\text{rc}$ produced by the refinement network $\Phi_\text{RF}$ closely matches the input in the facial region and preserves fine-grained details. As shown in the red bboxes in Fig.~\ref{suppl_fig9:refinement}, the base composite image $I_\text{base}$ produces boundary artifacts and does not blend the appearance smoothly across regions. In contrast, the refined image $I_\text{rc}$ mitigates boundary artifacts and achieves seamless blending across regions. The refinement network $\Phi_\text{RF}$ takes the base composite image $I_\text{base}$ as input, rather than the coarse complete image $I_\text{cc}$.

%===========================================================
%% Sec 8.3
\subsection{Without Prompt Conditioning}

Existing amodal completion~\cite{ozguroglu2024pix2gestalt,xu2024pdmc,liu2024ols_do} and HAC~\cite{noh2025sdhdo,zhou2021human_do} approaches are not designed to take user prompts as input. Consequently, although the proposed method accepts simple, user-friendly prompts, it is restrictive to assume that they are always available. To reduce prompt dependency, we apply prompt dropout during training, where the prompt image $I_\text{p}$ is replaced with an all-zero image with a probability of 5\%. With this training strategy, our method can perform HAC without a user prompt.

As shown in Table~\ref{suppl_table8:zero_prompt}, performance without a user prompt is slightly worse across all metrics than in the 2D pose-prompted setting, with the largest degradation observed in joint error. Nevertheless, our approach achieves the best performance on most metrics compared with previous methods that use a 2D pose map as the user prompt. As Table~\ref{suppl_table8:zero_prompt} is nearly identical to Table~\ref{table1:PHAC} of the main paper, we do not repeat the PGPIS results in this table. The detailed results for the PGPIS methods are reported in Table~\ref{table1:PHAC} of the main paper.

% Table 9: visible region
\begin{table}
    \centering
    \scriptsize
    \renewcommand{\arraystretch}{1.2}
    \setlength{\tabcolsep}{3pt}
    \resizebox{\columnwidth}{!}{
    \begin{tabular}{l|cccc|cccc}
        \toprule
        \raisebox{-2ex}{Method}
        & \multicolumn{4}{c|}{OccThuman2.0}
        & \multicolumn{4}{c}{AHP} \\
        \cline{2-5} \cline{6-9}
        & LPIPS* $\downarrow$ & SSIM $\uparrow$ & MSE* $\downarrow$ & PSNR $\uparrow$
        & LPIPS* $\downarrow$ & SSIM $\uparrow$ & MSE* $\downarrow$ & PSNR $\uparrow$ \\
        \midrule
        PIDM~\cite{bhunia2023pidm} & 88.43 & 0.862 & 16.16 & 19.31 & 120.85 & 0.835 & 20.40 & 17.51 \\
        MCLD~\cite{liu2025mcld} & 76.77 & 0.894 & 11.48 & 20.73 & 101.23 & 0.880 & 11.82 & 20.05 \\
        pix2gestalt~\cite{ozguroglu2024pix2gestalt} & 48.49 & 0.976 & 1.37 & 29.05 & 52.81 & 0.971 & 1.55 & 28.55 \\
        SDHDO~\cite{noh2025sdhdo} & \underline{38.08} & \underline{0.996} & \underline{0.32} & \underline{35.23} & \underline{37.71} & \underline{0.994} & \underline{0.32} & \underline{35.27} \\
        Ours & \textbf{13.72} & \textbf{0.998} & \textbf{0.17} & \textbf{37.93} & \textbf{15.65} & \textbf{0.998} & \textbf{0.14} & \textbf{38.85} \\
        \bottomrule
    \end{tabular}
    }
    \vspace{-2mm}
    \caption{\textbf{Quantitative comparison for PHAC in the visible region.} We compute LPIPS, SSIM, MSE, and PSNR on the visible (non-occluded) regions of the input image to measure fidelity to the input.}
    \vspace{-2mm}
    \label{suppl_table9:visible}
\end{table}

% Table 10: invisible region
\begin{table}
    \centering
    \scriptsize
    \renewcommand{\arraystretch}{1.2}
    \setlength{\tabcolsep}{3pt}
    \resizebox{\columnwidth}{!}{
    \begin{tabular}{l|cccc|cccc}
        \toprule
        \raisebox{-2ex}{Method}
        & \multicolumn{4}{c|}{OccThuman2.0}
        & \multicolumn{4}{c}{AHP} \\
        \cline{2-5} \cline{6-9}
        & LPIPS* $\downarrow$ & SSIM $\uparrow$ & MSE* $\downarrow$ & PSNR $\uparrow$
        & LPIPS* $\downarrow$ & SSIM $\uparrow$ & MSE* $\downarrow$ & PSNR $\uparrow$ \\
        \midrule
        PIDM~\cite{bhunia2023pidm} & 51.99 & 0.908 & 11.50 & 21.37 & 27.44 & 0.966 & 4.92 & 26.03 \\
        MCLD~\cite{liu2025mcld} & 47.67 & 0.927 & 7.65 & 23.25 & \underline{23.94} & \textbf{0.974} & \underline{3.34} & 27.24 \\
        pix2gestalt~\cite{ozguroglu2024pix2gestalt} & \underline{45.79} & \underline{0.932} & \underline{6.36} & \underline{24.32} & 23.96 & 0.972 & 3.67 & \underline{27.43} \\
        SDHDO~\cite{noh2025sdhdo} & 47.14 & 0.926 & 6.90 & 24.22 & 27.57 & 0.960 & 5.73 & 25.39 \\
        Ours & \textbf{36.85} & \textbf{0.950} & \textbf{4.29} & \textbf{26.15} & \textbf{23.05} & \underline{0.973} & \textbf{3.28} & \textbf{27.60} \\
        \bottomrule
    \end{tabular}
    }
    \vspace{-2mm}
    \caption{\textbf{Quantitative comparison for PHAC in the invisible region.} Using the same metrics as Table~\ref{suppl_table9:visible}, we compute them on the invisible (occluded) regions to measure reconstruction fidelity, conditioned on the visible appearance of the input image.}
    \vspace{-2mm}
    \label{suppl_table10:invisible}
\end{table}

%% Fig 10: additional qualitative results
\begin{figure*}
    \centering
    \includegraphics[width=1.0\linewidth]{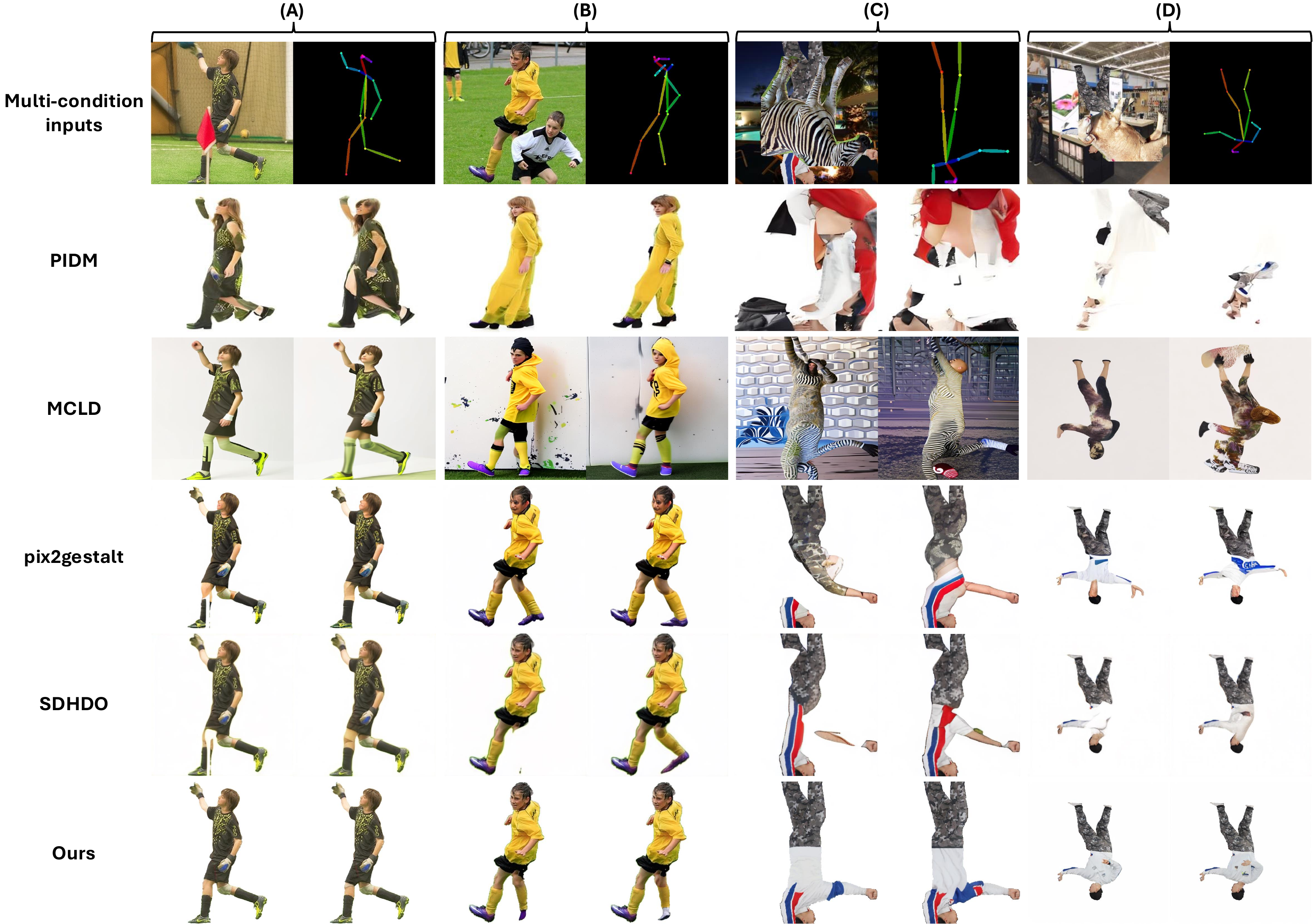}
    \vspace{-6mm}
    \caption{\textbf{Additional qualitative comparison for PHAC.} (A, B): AHP test examples, where previous methods either hallucinate appearance, lose fine-grained details, or fail to complete occluded regions. In contrast, our method preserves the visible appearance and aligns with the pose condition. (C, D): OccThuman2.0 test examples under severe occlusions, where previous methods often produce implausible or pose-misaligned results. In contrast, our method synthesizes occluded regions that remain consistent with both the visible appearance and the pose condition.}
    \label{suppl_fig10:add_qual}
    \vspace{-2mm}
\end{figure*}

%===========================================================
%% Sec 8.4
\subsection{Region-wise Quantitative Results}

Table~\ref{table1:PHAC} of the main paper reports quantitative results over the entire image, covering both visible and invisible regions. We additionally report region-wise quantitative results for the visible and invisible regions in Tables~\ref{suppl_table9:visible} and~\ref{suppl_table10:invisible}, respectively. Since we evaluate the visible and invisible regions separately, whole-image metrics such as KID and joint error are not applicable. We therefore report only reconstruction metrics (MSE, PSNR) and perceptual quality metrics (LPIPS, SSIM).

As shown in Table~\ref{suppl_table9:visible}, PIDM and MCLD tend to overfit the training set, failing to preserve the visible appearance and resulting in weak performance in the visible region across datasets. Amodal completion methods better preserve the visible regions than PGPIS methods, with SDHDO substantially outperforming pix2gestalt. With the refinement network, our method further improves performance and outperforms all previous methods on all metrics.

As shown in Table~\ref{suppl_table10:invisible}, the gap between PGPIS and amodal completion methods is comparatively modest in the invisible region, in contrast to the larger gap observed in Table~\ref{suppl_table9:visible}. On OccThuman2.0, we obtain the best overall performance, with a clear margin over the top three competing methods~\cite{liu2025mcld,ozguroglu2024pix2gestalt,noh2025sdhdo}, while on AHP our approach leads on three of the four metrics.

%===========================================================
%% Sec 8.5
\subsection{Additional Qualitative Results}
\label{suppl_sec8.5:add_qual_results}

\begin{figure}
    \centering
    \includegraphics[width=0.9\linewidth]{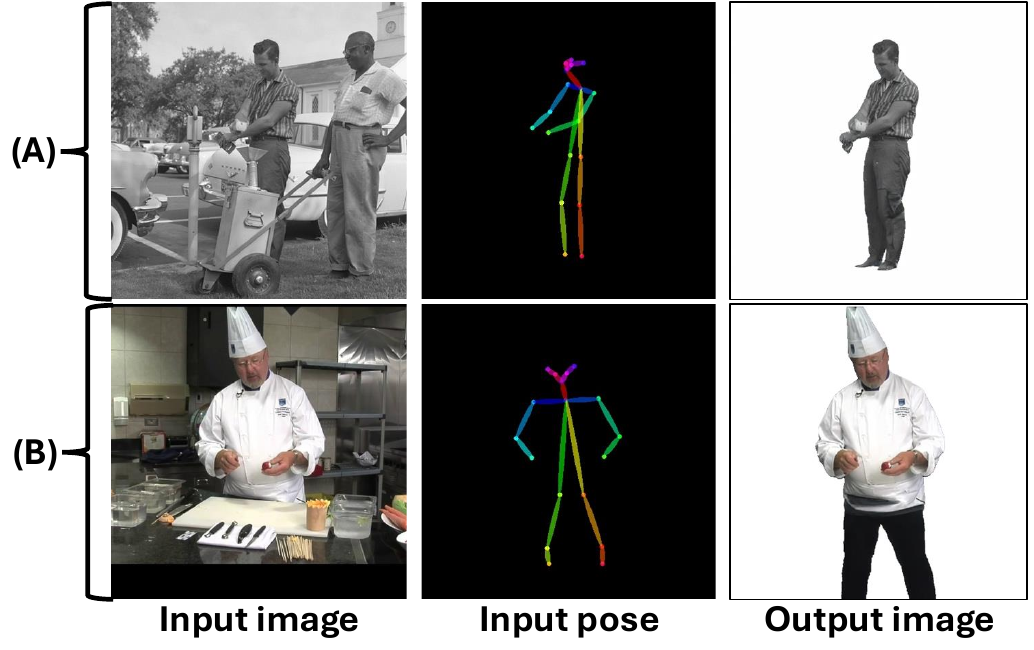}
    \vspace{-2mm}
    \caption{\textbf{In-the-wild qualitative results.} We show in-the-wild results of our method on a grayscale MSCOCO example (A) and an unseen-identity MPII example (B).}
    \label{suppl_fig11:itw}
    \vspace{-2mm}
\end{figure}

\begin{figure}
    \centering
    \includegraphics[width=0.9\linewidth]{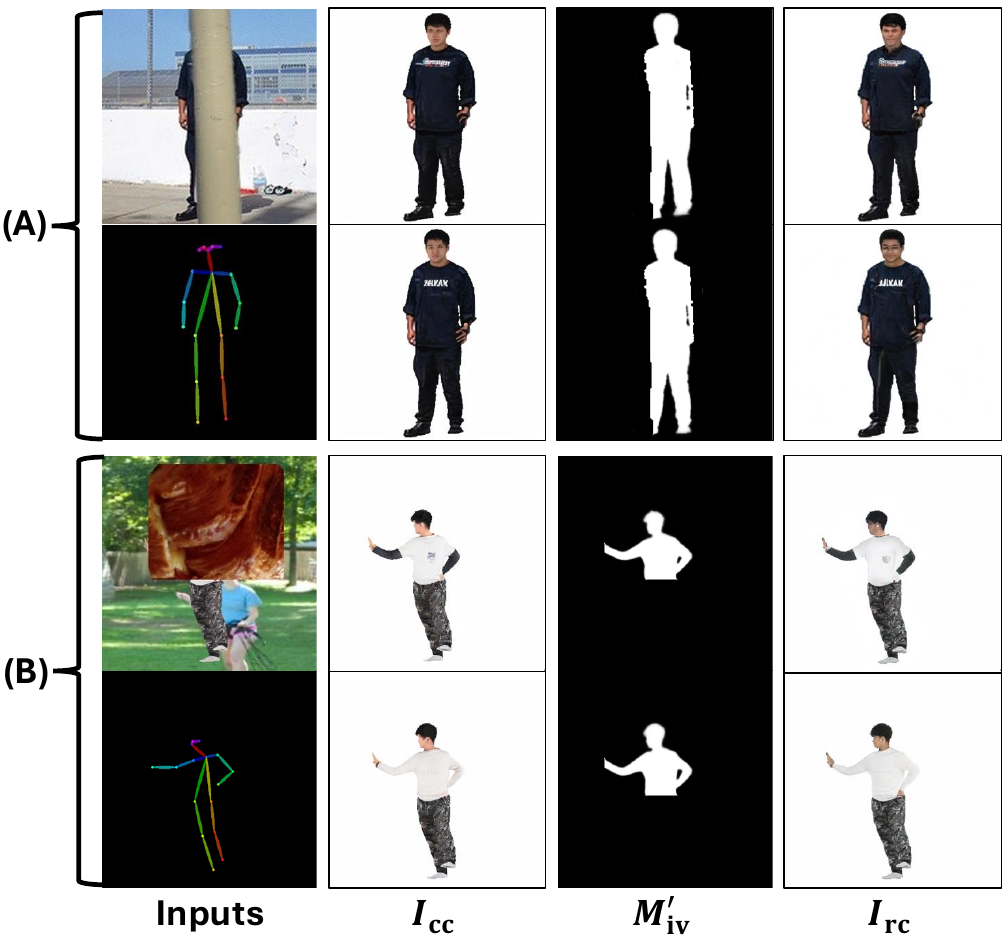}
    \vspace{-2mm}
    \caption{\textbf{Severe-occlusion qualitative results.} We show intermediate outputs ($I_\text{cc}$, $M_\text{iv}'$) and the final output ($I_\text{rc}$) of our method under severe occlusions. (A): AHP real example (76\% occlusion). (B): OccThuman2.0 example (53\% occlusion). Each row uses a different random seed.}
    \label{suppl_fig12:severe_occ}
    \vspace{-2mm}
\end{figure}

In this section, we present additional qualitative results for PHAC in Figs.~\ref{suppl_fig10:add_qual}--\ref{suppl_fig12:severe_occ}. Fig.~\ref{suppl_fig10:add_qual} shows results on the AHP test dataset (A, B) and the OccThuman2.0 test dataset (C, D). Fig.~\ref{suppl_fig11:itw} presents in-the-wild results on MSCOCO~\cite{lin2014microsoft_coco} and MPII~\cite{andriluka2014mpii}, and Fig.~\ref{suppl_fig12:severe_occ} provides additional examples under severe occlusions. As described in Sec.~\ref{suppl_sec7.4:prompt_image}, we construct the prompt images from the multi-condition inputs in Fig.~\ref{suppl_fig10:add_qual}, and all baselines are conditioned on the same 2D pose map except that MCLD uses a DensePose UV map and pix2gestalt does not use an explicit conditioning input.

As shown in Fig.~\ref{suppl_fig10:add_qual}(A, B), PIDM and MCLD hallucinate appearance rather than preserving the visible appearance in the input image. In contrast, pix2gestalt often fails to complete occluded regions (Fig.~\ref{suppl_fig10:add_qual}(A)) and generates images that lack fine-grained detail, particularly in the facial region (Fig.~\ref{suppl_fig10:add_qual}(B)). SDHDO preserves the visible appearance more faithfully, but produces blurry images and may still fail to reconstruct occluded regions, similar to pix2gestalt. Overall, our method consistently produces images that align with the pose condition while preserving the visible appearance.

Fig.~\ref{suppl_fig10:add_qual}(C, D) presents OccThuman2.0 results under more severe occlusions than (A, B). Under heavy occlusions, PIDM often fails to produce a plausible person image. Due to errors in UV map estimation, MCLD may transfer occluder appearance into the output and produce implausible results. SDHDO and pix2gestalt yield either implausible (Fig.~\ref{suppl_fig10:add_qual}(C)) or pose-misaligned outputs (Fig.~\ref{suppl_fig10:add_qual}(D)). In contrast, our method reconstructs occluded regions consistent with the visible appearance and closely aligns the synthesized content with the pose condition, producing plausible person images even under severe occlusions. For example, in Fig.~\ref{suppl_fig10:add_qual}(D), other methods fail to reconstruct the right hand correctly, whereas our method synthesizes it consistently with the pose condition.

%% Fig 11: in-the-wild qualitative results

%% Fig 12: severe occlusion qualitative results

Fig.~\ref{suppl_fig11:itw} reports in-the-wild results of our method, where (A) shows a grayscale MSCOCO example and (B) shows an unseen-identity MPII example. These results demonstrate that our method can perform PHAC on in-the-wild data while preserving the visible appearance of the input image and remaining consistent with the input pose.

We additionally provide qualitative results under severe occlusions in Fig.~\ref{suppl_fig12:severe_occ}. For each example, we visualize the multi-condition inputs as well as the intermediate and final outputs of our method: the coarse complete image $I_\text{cc}$, the dilated invisible mask $M_\text{iv}'$, and the refined complete image $I_\text{rc}$. Fig.~\ref{suppl_fig12:severe_occ}(A) shows an AHP real example where approximately 76\% of the human body is occluded, and Fig.~\ref{suppl_fig12:severe_occ}(B) shows an OccThuman2.0 example with approximately 53\% occlusion. Each row corresponds to a different random seed during inference. Even in these severe-occlusion cases, our method robustly aligns with the input pose and produces plausible $I_\text{cc}$, $M_\text{iv}'$, and $I_\text{rc}$. Due to the limited visible-appearance cues under heavy occlusion, varying the seed can lead to different texture realizations in the invisible regions, while maintaining pose consistency. For reference, the OccThuman2.0 examples in Fig.~\ref{suppl_fig10:add_qual}(C, D) correspond to approximately 58\% and 43\% occlusion, respectively.

%% Table 11: dilation ablation
\begin{table}
  \centering
  {\scriptsize
  \begin{tabular}{l|cccccc}
    \toprule
    & LPIPS* $\downarrow$ & KID* $\downarrow$ & MSE* $\downarrow$ & PSNR $\uparrow$ & Joint Err. $\downarrow$ \\
    \midrule
    w/o dil & 49.86 & 6.54 & 4.59 & 25.63 & 23.73 \\
    w/ dil  & \textbf{49.47} & \textbf{6.12} & \textbf{4.37} & \textbf{25.86} & \textbf{23.33} \\
    \bottomrule
  \end{tabular}
  }
  \vspace{-2mm}  
  \caption{\textbf{Quantitative comparison of our method with and without dilation.} Invisible mask dilation is used as a conservative margin for boundary uncertainty and has a marginal impact on metrics.}
  \vspace{-2mm}
  \label{suppl_table11:dilation}
\end{table}

%===========================================================
%% Sec 8.6
\subsection{Effect of Invisible Mask Dilation}

As discussed in Sec.~\ref{sec3.4:refine}, we use invisible mask dilation only as a conservative margin to account for boundary uncertainty in the predicted invisible region. Table~\ref{suppl_table11:dilation} reports an ablation comparing our method with and without dilation on the OccThuman2.0 test dataset. Dilation yields consistent but marginal improvements across metrics, indicating that it has limited impact on overall performance while providing a safer margin around uncertain occlusion boundaries.

\clearpage

{
    \small
    \bibliographystyle{ieeenat_fullname}
    \bibliography{main}
}

\end{document}